\documentclass[journal]{IEEEtran}
\usepackage{times}
\usepackage{epsfig}
\usepackage{graphicx}
\usepackage{amsmath}
\usepackage{amssymb}
\usepackage{authblk}

\usepackage{cite}

\ifCLASSINFOpdf
\else
\fi
\begin{document}
%
\title{Large-scale, Fast and Accurate Shot Boundary Detection through Spatio-temporal Convolutional Neural Networks}
%
%
%


\author[1]{Ahmed Hassanien}
\author[1]{Mohamed Elgharib}
\author[2]{Ahmed Selim}
\author[3]{Sung-Ho Bae}
\author[4]{Mohamed Hefeeda}
\author[3]{Wojciech Matusik}
\affil[1]{Qatar Computing Research Institute, HBKU} \affil[2]{Trinity College Dublin, CONNECT Center} \affil[3]{MIT CSAIL} \affil[4]{Simon Fraser University} 

\maketitle

\begin{abstract}
	Shot boundary detection (SBD) is an important pre-processing step for video manipulation. Here, each segment of frames is classified as either sharp, gradual or no transition. Current SBD techniques analyze hand-crafted features and attempt to optimize both detection accuracy and processing speed. However, the heavy computations of optical flow prevents this from happening. To achieve this aim, we present an SBD technique based on spatio-temporal Convolutional Neural Networks (CNN). Since current datasets are not large enough to train an accurate SBD CNN, we are the first to present a very large SBD dataset that allows deep neural networks techniques to be effectively applied. Our dataset contains more than 3.5 million frames of sharp and gradual transitions. The transitions are generated synthetically using image compositing models. Our dataset contain additional 70,000 frames of important hard-negative no transitions. We perform the largest evaluation to date for one SBD algorithm, on real and synthetic data, containing more than 4.85 million frames. In comparison to the state of the art, we outperform dissolve gradual detection, generate competitive performance for sharp detections and produce significant improvement in wipes. In addition, we are up to 11 times faster than the state of the art.  
\end{abstract}


%
\IEEEpeerreviewmaketitle

\begin{figure*}
  \centering
   \includegraphics[height=6cm, width=16cm]{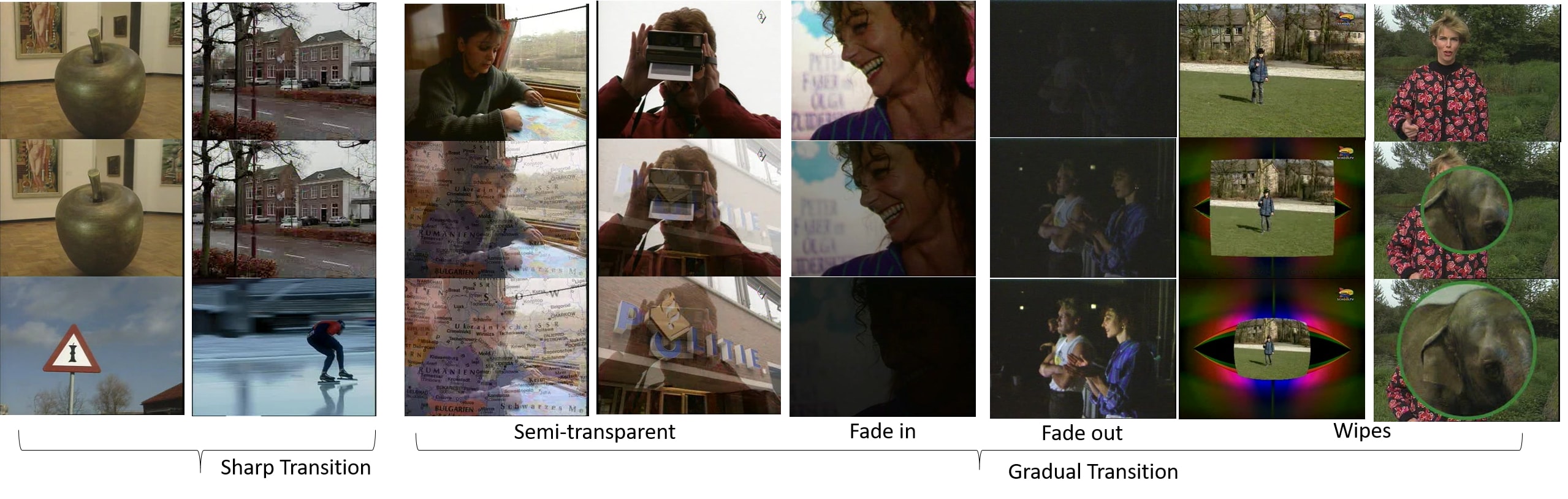}
   \caption{Shot transitions are classified into two main categories: sharp and gradual. Gradual transitions are further classified into soft and wipes. Soft include semi-transparent, fade in and fade out. Wipes are the most ill defined form of transitions.}
\label{fig:Intro}
 \end{figure*}

\section{Introduction}

With the wide adoption of digital video, the demand for editing and manipulating video content is in continuous rise. This, however, requires better understanding of videos and their composition. Videos are composed of different camera shots placed after each other. A video shot transitions into another through several forms of visual effect. These visual effects can be classified into two main categories: sharp and gradual \cite{Smeaton10} as shown in Figure~\ref{fig:Intro}. The former is a sudden change of the shot over 1 frame, while gradual transitions occur over multiple frames. Gradual transitions are further classified into dissolve and non-dissolve. The former includes cases such as semi-transparent graduals, fade in and fade out (see Figure~\ref{fig:Intro}). Non-dissolve are dominated by wipes (see Figure~\ref{fig:Intro}). Wipe graduals have a much wider variety than the dissolve graduals. 

\begin{figure}
  \centering
   \includegraphics[height=7cm, width=7cm]{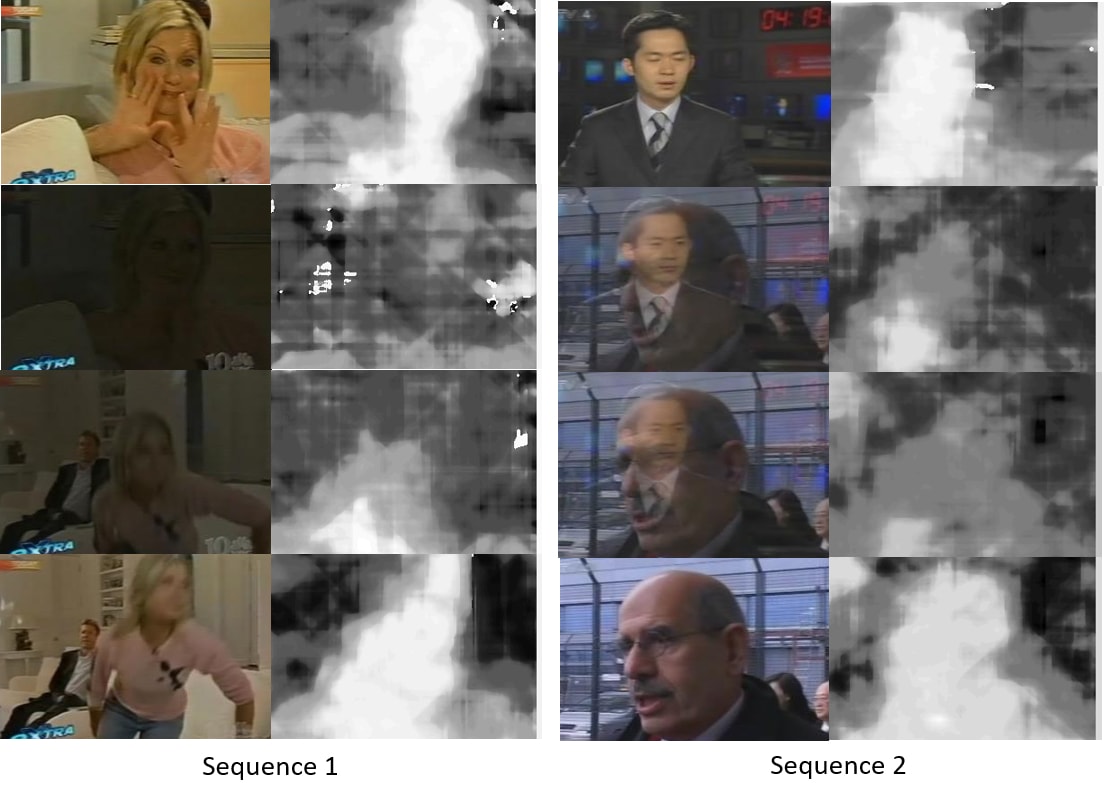}
   \caption{A figure illustrating the importance of shot boundary detection during 2D-to-3D conversion. Here, we show consecutive frames from two sequences. For each frame we show its original RGB image (left) and the corresponding disparity map (right) as estimated by the 2D-to-3D conversion technique of Bae et al. \cite{Bae17}. The disparity maps is a gray-scale image where white represents objects close to the screen. Note how the disparity captures the humans outlines before and after transitions (first and last frame). During gradual transition, however, the outline is severely destroyed. This is because gradual transitions violate the single layer model used by many video processing techniques. Hence, it is desirable to detect such gradual transitions and to give them a special treatment during post-processing. For this, shot boundary detection is essential.}
\label{fig:DeepView}
 \end{figure}

Video post-processing techniques are in rising popularity and they cover a wide range of applications. This includes video coding \cite{Fan00}, visual quality enhancement \cite{Xie16b,Kim16b}, graphics rendering \cite{Xie16a,Calagari15,Bae17}, video understanding \cite{Zhang16a,Zhang16b} and many others \cite{Song16,Templin14}. Such post-processing techniques, however, are based on assumptions, some of which can be violated during shot transitions. For instance, many techniques assume the presence of one layer at one spatial point, an assumption heavily violated during dissolve transitions. This can lead to unpleasant artifacts as in the case of 2D-to-3D conversion (see Figure~\ref{fig:DeepView}). Here, the disparity maps can undergo strong artifacts during gradual transitions. Hence, detecting video transitions and assigning a special treatment for them during post-production is an important and desirable step. However, with the high computational demand of many post-production techniques, as well as the real-time requirement of some, shot boundaries detection (SBD) needs to be performed with both high detection accuracy and very fast processing speed.

Current SBD techniques analyze hand-crafted features \cite{Priya14,Lu13,Smeaton10,Mohanta12,Adjeroh09,Cernekova06,Lankinen13,Lelescu03,Zhang12,Thounaojam16}. Fast techniques analyze only spatial information such as intensity histogram \cite{Lu13,Zhang12}, edges \cite{Adjeroh09}, mutual information and others \cite{Cernekova06,Thounaojam16,Lelescu03,Lankinen13}. Such techniques, while being fast, generate poor detection. To boost detection, motion information is incorporated through optical flow \cite{Priya14,Smeaton10,Yuan04,Liu07}. However, the heavy computations of optical flow \cite{Tao12,Dosovitskiy15,Baker11} make such techniques slow. As SBD techniques are commonly used as a pre-processing step for video manipulation, optimizing both their detection accuracy and processing speed is important. This, however, remains a challenging problem.   


We present DeepSBD, a fast and accurate shot boundary detection through convolutional neural networks (CNN). We exploit big data to achieve high detection performance. In addition, we exploit the parallelizable nature and common GPU implementations of CNNs to achieve fast processing speed. Our technique takes a segment of 16 frames as input, and classifies it as either gradual, sharp or no-transition. It analysis both spatial and temporal information through an effective 3D convolutional network for video processing, inspired by C3D \cite{Tran15}. 

To train our network, we need a well-annotated very large dataset. Despite datasets already exist from the TRECVID challenge and others \cite{Smeaton10,Baraldi15}, experiments show they are not sufficient to train a high accuracy CNN solution. In addition, the vast majority of these datasets are used for testing and evaluating different techniques, and hence should not be used for training. To overcome this problem, we present a very large SBD dataset with clean and accurate annotations capable of training a highly accurate CNN SBD solution.    
This also allows us to test on all available TRECVID data (3.9 million frames) \cite{Smeaton10}. The first dataset portion, SBD\_Syn, is generated synthetically using image compositing models. It contains 220,339 sharp and gradual segments, each segment contains 16 frames. The second portion, SBD\_BT, contains 4,427 no transition segments. They are carefully manually annotated in a way to improve detector's precision; they act as hard-negatives. We optionally use 1 TRECVID release (2005) and another SBD dataset of Baraldi et al. \cite{Baraldi15} to further improve performance. These datasets have 18,027 total transitions with prior annotations. That is only 7\% of all training datasets.   
%

Aspects of novelty of our work include:
\begin{enumerate}
\item The first CNN SBD technique. We outperform dissolve gradual detection, generate competitive performance for sharp detections and produce significant improvement in wipes. In addition, we are up to 11 times faster than the state of the art.
\item Introduction of a new very large SBD dataset for training an accurate CNN model. Our dataset contains 3.5 million frames of synthetic transitions and 70,000 frames of hard negative no-transitions.
\item A large wipes dataset containing 1.1 million frames. We will release all our data-sets and code to encourage future research.
\item The largest SBD evaluation to date on 4.85 million frames. 3.9 million frames are from all TRECVID years \cite{Smeaton10} while most of the rest are synthetically generated.  
\end{enumerate}

The next section reviews the state of the art. Here, we discuss the main components of our solution including current SBD techniques, current available SBD datasets and CNN solutions for video spatio-temporal analysis. We then present our SBD solution with emphasize on our detection system and our dataset generation process. Section IV presents detailed results and analysis. The results are also supported by a supplementary material (in .pdf format, please examine). Section V is conclusion.   



\section{State of the Art}

\subsection{Shot Boundary Detection Techniques}

SBD techniques \cite{Smeaton10,Priya14,Lu13,Liu07} extract features and analyze them temporally. Detection is then performed by finding temporal profiles that fit the examined transition model. Sharp transitions undergo a sudden change in the temporal profile over one frame. Gradual transitions exhibit a more stretched change in time. Current SBD techniques are classified into two main categories: spatial-only and spatio-temporal analysis based. The former estimates the temporal profile by comparing only spatial features \cite{Priya14,Lu13,Smeaton10,Mohanta12,Adjeroh09,Cernekova06,Lankinen13,Lelescu03,Zhang12,Thounaojam16}. A number of spatial features are used such as color histograms \cite{Lu13,Zhang12}, edges \cite{Adjeroh09}, mutual information and Entropy \cite{Cernekova06}, wavelet representations \cite{Priya14}, SURF \cite{Apostolidis14} and many others \cite{Thounaojam16,Lelescu03,Lankinen13,Berladi15}. 

Spatial-only SBD methods generate conservative detection accuracy with fast processing speed. Spatio-temporal techniques use optical flow to make detection more robust to scene and camera motions \cite{Priya14,Mohanta12,Lian11,Kawai07,Kawai07}. Such motions can arise due to camera movements and shakiness and often confuse the detection process. Hence, optical flow \cite{Tao12,Dosovitskiy15,Baker11} between neighboring frames is estimated and removed through frame interpolation. Analysis of the temporal profile is then proceeded as in the spatial-only techniques. Here, motion compensation often reduces false detections of SBD. The main drawback of spatio-temporal techniques, however, is the heavy computations of optical flow.     

Among the rich SBD literature, four of the best performing and/or most recent techniques are Liu et al. \cite{Liu07}, Yuan et al. \cite{Yuan05}, Lu et al. \cite{Lu13} and Priya et al. \cite{Priya14}. Lu et al. focuses more on generating fast results and hence they do not incorporate motion information. Their technique is based on assessing temporal discontinuities through HSV histogram. 
Priya et al. \cite{Priya14} proposed a wavelet based feature vector that measures four main quantities: color, edge, texture, and motion strength. 
The feature vector is extracted for each frame of a sequence and the temporal profile is estimated through frame differencing. Liu et al. \cite{Liu07} uses a large number of features including color, histogram, edge, motion and related statistical features. Liu et al., Priya et al. and Yuan et al. all focus on high detection accuracy. This, however, comes with the high cost of optical flow. Furthermore, the techniques of Apostolidis et al. \cite{Apostolidis14} and Berladi et al. \cite{Berladi15} were recently released. They analyze only spatial information such as SUFR and HSV/color histogram and hence often generate conservative performance with fast processing speed.  

To the best of our knowledge, Liu et al. \cite{Liu07} is the latest wipe detector. A candidate transition segment is proposed and the difference between each frame and the start and end frame is calculated. This generates two curves, one for the start and another for the end of the segment. For wipes, the curves should have opposing gradients and somewhat linear. Furthermore, to reduce errors due to camera and object movements, motion compensated frame differencing is used. 



%

\subsection{SBD Datasets}

Between the years 2001 to 2007, the National Institute of Standards and Technology (NIST) \cite{NIST} maintained data for the TRECVID shot boundary detection (SBD) challenge \cite{Smeaton10}. The dataset contains a wide variety of content including color, gray-scale, indoor, outdoor, outer-space and different levels of noise. The dataset has a total of 4,333,153 frames with 24,423 transitions, $64\%$ of which are sharp. The rest are gradual. Transitions were manually annotated in a way to distinguish between sharp and graduals. 
Four more releases from a different challenge were maintained by NIST that contain data relevant to SBD. The releases are T2007t, T2007d, T2008 and T2009, containing 34,765,424 frames with 155,902 transitions. The annotations of these data, however, do not distinguish between sharp and graduals. 
Finally, one more data release related to SBD was generated by Baraldi et al. \cite{Baraldi15}. Here, the authors addressed the different application of video scene segmentation.



We collected all the SBD related dataset. However, some TRECVID data appear not to exist anymore and/or they can not be tracked. Despite being a large dataset, several factors prevent them to be used for training. First, most of T2001 and T2002 should be removed due to their poor annotations. In addition, at least T2007 and the rest of T2001 should be removed as they are commonly used for evaluation \cite{Priya14,Lu13,Liu07}. This leaves at most 15,163 sharp and 7,274 gradual annotations from TRECVID and Baraldi et al. \cite{Baraldi15}. Experiments show this is not sufficient to train an accurate SBD CNN. 

\subsection{Spatio-temporal analysis using CNN} 


Our solution analyzes both spatial and temporal information through CNN. Hence, our network is related to the literature on video classification. Karpathy et al. \cite{Karpathy14} proposed multiple approaches for extending the connectivity of CNN to take advantage of the spatio-temporal information. Results show that CNN can generate strong improvement over hand-crafted features. However, the multiple frame models showed a modest improvement compared to the single-frame model. Next, Simonyan et al. \cite{Simonyan14} proposed a two stream CNN network for video classification. One network analyzes the spatial information while the second analyzes the optical flow field. Their approach generates significant improvement over the single frame model of \cite{Karpathy14}.     

Tran et al. \cite{Tran15} presented the first single stream CNN that incorporate both spatial and temporal information at once. Their approach takes multiple frames as input and examines them with 3D spatio-temporal convolutional filters. They handle the problem of activity recognition and performed evaluation on the UCF101 dataset. They outperformed all previous work, including \cite{Karpathy14,Simonyan14}. In addition, their technique is fast as does not require optical flow estimation. 

Our solution is a full Shot Boundary Detection (SBD) system consisting of a CNN-based classification step, a merging step and a post-processing step. At the core of our CNN-classification is a spatio-temporal architecture inspired by Tran et al. \cite{Tran15}. Unlike Tran et al. \cite{Tran15}, however, our architecture uses batch normalization. Furthermore, our solution contains a component for generating very large well annotated data-sets for training our SBD. Results show that all components of our solution, including dataset generation and our full SBD system, play an important role in outperforming the state of the art, both in detection accuracy and processing speed.


\section{Our Approach}


\subsection{Algorithm Design}

\begin{figure*}
  \centering
   \includegraphics[height=4cm, width=18cm]{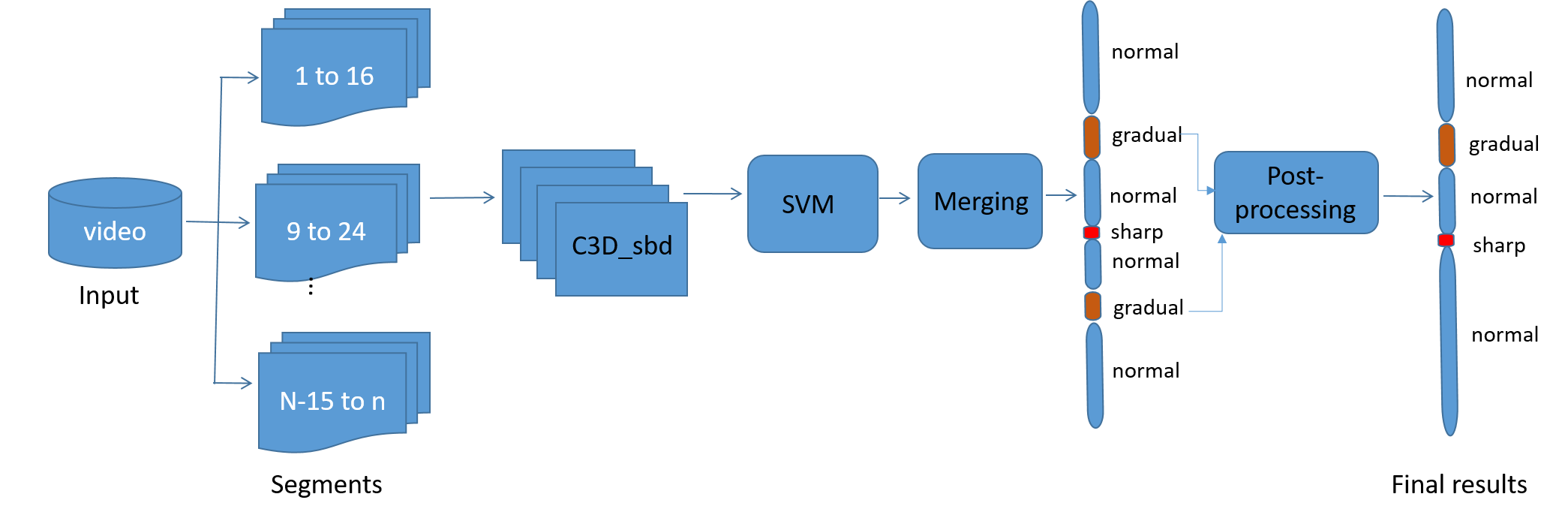}
   \caption{Overview of our Shot Boundary Detection (SBD) system. A video is divided into segments of 16 frames with an overlap of 8. Each segment is fed to a 3D CNN (see Table~\ref{tab:DeepSBD_Arch}). The output of fc8 is fed to an SVM and labels are assigned. Consecutive segments with the same labeling are merged. Finally, false alarms of gradual transitions are reduced through a histogram-driven temporal differencing. }
\label{fig:DeepSBD}
 \end{figure*}

\begin{figure*}
  \centering
   \includegraphics[height=2.5cm, width=18cm]{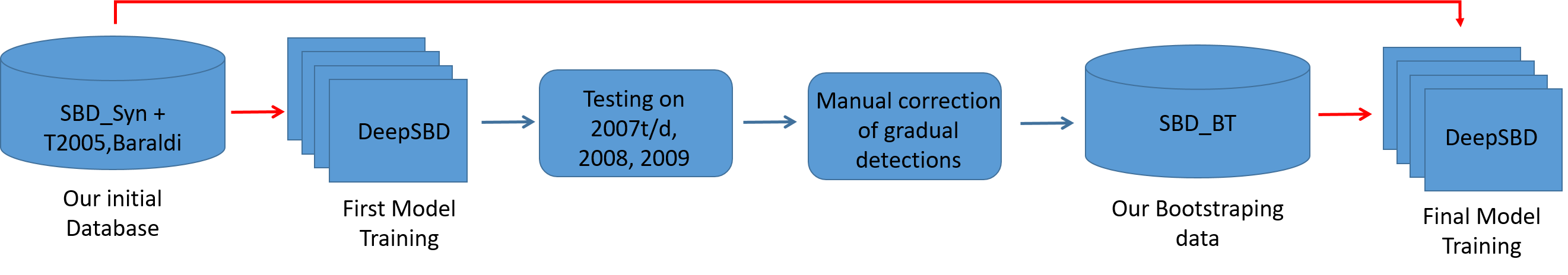}
   \caption{Our process of data generation. Here, red arrows refer to data fed to CNN training. We first use our synthetic data SBD\_Syn and some real data (T2005 and Baraldi et al.) to train our solution. The model runs on 2007t/d, 2008 and 2009. We manually correct gradual detections into sharp, gradual and no transitions. This generates our bootstrapping data SBD\_BT. We train the final solution using both SBD\_Syn and SBD\_BT, and optionally T2005 and Baraldi et al. .}
\label{fig:SBDData}
 \end{figure*}

We present a technique for automatic detection and classification of shot boundaries. We name our technique DeepSBD. A video is divided into segment of frames. Each segment is assigned one of three labels: 1) sharp transition, 2) gradual transition or 3) no transition. We use segments of length 16, with an overlap of 8. Each segment is fed to a deep 3D-CNN that analysis both spatial and temporal information. Our network, C3D\_{sbd}, is inspired by \cite{Tran15} and is trained from scratch for shot boundary detection. The last feature layer is fed to an SVM classifier. This gives the first labeling estimate. Consecutive segments with the same labeling are merged and the result is passed to a post-processing step. The step reduces false positives with little motion. For such segments, we estimate the color histogram of the first and end frame. We measure the Bhattacharyya distance between these histograms. If the distance is small, we declare this segment as no-transition. We use an OpenCV implementation for both color histogram and Bhattacharyya distance, which is very fast.

Figure~\ref{fig:DeepSBD} shows an overview of our detection system. Our network, C3D\_{sbd}, consists of five 3D convolutional layers (see Table~\ref{tab:DeepSBD_Arch}). All convolutional layers are followed by Rectified Linear Unit (ReLU) and pooling layers. The first two convolutional layers are followed by Local Response Normalization (LRN). Two fully connected layers exist, fc6 and fc7, each containing 2049 neurons. The last fully connected layer fc8 contain only 3 neurons, one for each class (sharp, gradual and no transition). In comparison to \cite{Tran15}, C3D\_{sbd} uses batch normalization after the first two convolutional layers. 

\begin{table}
\small
\begin{tabular}{|c|c|c|c|}
\hline
\textbf{Layer}   & \textbf{Kernel} & \textbf{Feature map} & \textbf{Followed} \\ 
                 & (t$\times$y$\times$x$\times$c)$\times$f             &    \textbf{dimension}  & \textbf{    by   }                      \\ \hline
Data    & -                           & 20$\times$3$\times$16$\times$112$\times$112                 &                     \\ \hline
Conv1   & (3$\times$3$\times$3$\times$3)$\times$96                & 20$\times$96$\times$14$\times$55$\times$55                  & ReLU                \\ 
        &                             &                                 & LRN                 \\ \hline
Pool1   & -                           & 20$\times$96$\times$12$\times$27$\times$27                  &                     \\ \hline
Conv2   & (3$\times$3$\times$3$\times$96)$\times$256              & 20$\times$256$\times$12$\times$29$\times$29                 & ReLU                \\ 
        &                             &                                 & LRN                 \\ \hline
Pool2   & -                           & 20$\times$256$\times$10$\times$14$\times$14                 &                     \\ \hline
Conv3   & (3$\times$3$\times$3$\times$256)$\times$384             & 20$\times$384$\times$10$\times$14$\times$14                 & ReLU                \\ \hline
Conv4   & (3$\times$3$\times$3$\times$384)$\times$384             & 20$\times$384$\times$10$\times$14$\times$14                 & ReLU                \\ \hline
Conv5   & (3$\times$3$\times$3$\times$384)$\times$256             & 20$\times$256$\times$10$\times$14$\times$14                 & ReLU                \\ \hline
Pool5   & -                           & 20$\times$256$\times$8$\times$7$\times$7                    &                     \\ \hline
Fc6     & (8$\times$7$\times$7$\times$256)$\times$2048            & 20$\times$2048$\times$1$\times$1$\times$1                   & ReLU                 \\ 
        &                             &                                 & Drop 0.5                 \\ \hline

Fc7     & 2048$\times$2048                   & 20$\times$2048$\times$1$\times$1$\times$1                   & ReLU          \\ 
        &                             &                                 & Drop 0.5                 \\ \hline

Fc8     & 2048$\times$3                      & 20$\times$3$\times$1$\times$1$\times$1                      &                     \\ \hline
Softmax & Label                       & -                               &                     \\ \hline
\end{tabular}\vspace{3pt}
\caption{The model parameters of C3D\_{sbd}. In comparison to \cite{Tran15}, C3D\_{sbd} uses batch normalization. 
}
\label{tab:DeepSBD_Arch}
\end{table}

\subsection{Dataset Generation}\label{Sec:Dataset}

\begin{figure}
  \centering
   \includegraphics[height=3cm, width=7cm]{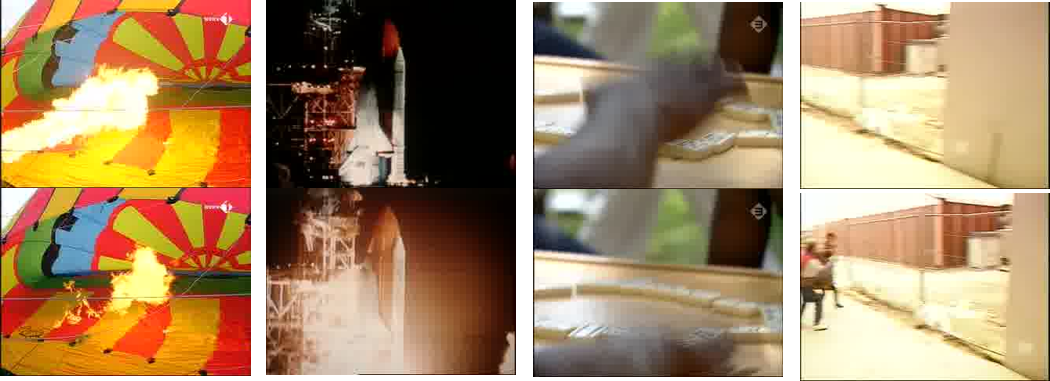}
   \caption{No-transition samples from our bootstrapping data (SBD\_BT). They contain complex cases such as fast motion, occlusion and illumination variation. These cases can be misclassified as graduals. More examples are in the supplementary material, Figure~2.}
\label{fig:BTData}
 \end{figure}

Training an SBD CNN requires a large and well-annotated dataset. We present two datasets, SBD\_Syn (Table~\ref{tab:DeepSBDSyn}) and SBD\_BT (Table~\ref{tab:DeepSBDBT}). SBD\_Syn is generated synthetically while SBD\_BT is generated in a way to improve detector's precision, through bootstrapping. Figure~\ref{fig:SBDData} shows the process of generating both datasets. We first use SBD\_Syn with T2005 and Baraldi et al. to train from scratch our solution. We run this solution on data from T2007t/d, T2008 and T2009. Due to the massive size of these datasets, however, we only examine segments originally annotated as any form of transition. Note that original annotations here do not distinguish between sharp or gradual. We closely examine segments detected as graduals. We manually filter them into three classes: gradual, sharp and no transitions. The no-transition represent complicated hard-negative cases such as illumination variation and fast motion (see Figure~\ref{fig:BTData}). Finally, we train from scratch a final solution using both SBD\_Syn and SBD\_BT. We optionally use T2005 and Baraldi et al. to further improve performance. 
Results show that SBD\_BT has a great impact in reducing false detections and improving the overall performance. The supplementary material shows images from the datasets of SBD\_Syn and SBD\_BT in Figure~1 and Figure~2.   

\textbf{SBD\_Syn:} 
Table~\ref{tab:DeepSBDSyn} shows the content of SBD\_Syn. Images from this dataset is shown in the supplementary material (Figure 1). 
The dataset is generated synthetically through image compositing models \cite{Levin07}. A transition is modeled as a linear combination between the underlying shots 
\begin{equation}
I_t({\bf {x}}) = \alpha_t({\bf {x}})B_t({\bf {x}}) + (1-\alpha_t({\bf {x}}))F_t({\bf {x}})
\label{eq:DataGen}
\end{equation}
Here, $I_t$ denotes the observed frame at time $t$, while $B$ and $F$ are the content from the previous and next shots respectively. $\alpha$ is the mixing parameter between both shots while ${\bf {x}}$ denotes image pixels. The values and distribution of $\alpha$ define the type of shot transition. If no transition exist, then $(\alpha_t,\alpha_{t+1})=({1,1})$. A sharp transition, however, have a sudden temporal change with $(\alpha_t,\alpha_{t+1})=(1,0)$. For gradual transitions, $\alpha$ changes over time from $1$ to $0$. This change occurs over a set of frames and hence $(\alpha_t,.....\alpha_t+N)=(1,.,.,1-t/N,.,0)$. $N$ is the transition duration and $t$ is the frame index where $t=0$ denotes the last frame of the previous shot. Here, the in-between $\alpha$ values are non-binary. This generates the dissolve nature of most gradual transitions (Figure~\ref{fig:Intro}). For wipes, $\alpha$ is spatially-varying aswell as temporally-varying. 

To generate SBD\_Syn we need to define $F$, $B$ and $\alpha$ in Eq.~\ref{eq:DataGen}. $F$ and $B$ must not contain any shot transitions. We use the T2007t/d, T2008, T2009 and their annotations to find such frames. We sample $F$ and $B$ in a way to ensure a large offset from the nearest transition. Sharp transitions are generated by applying Eq.~\ref{eq:DataGen} with $(\alpha_t,\alpha_{t+1})=(1,0)$. Gradual transition generation, however, is more complex. For SBD\_Syn we focus on dissolve gradual generation. We randomly select the transition duration $N$, where $N<=16$. We also randomly select the transition start and end frames for both $B$ and $F$. We draw $N$ $\alpha$ samples, where $\alpha$ is modeled with a uniform distribution. We sort all $\alpha$ values in descending order and apply Eq.~\ref{eq:DataGen} for each of the $N$ frames. 

We train C3D\_{sbd} using balanced data for sharp, gradual and no-transition. We experimented with different data sizes. We found 40,000 segments for each class generate good results. We also train the SVM for sharp and gradual using 110,000 segments for each. For CNN, we use a step learning policy. Learning rate starts with a value of 0.0001 and is reduced gradually by a factor of 10 every two epochs. We use a batch size of 20, and train the model for 6 epochs. That is two epochs for each learning rate of 1e-4, 1e-5, and 1e-6. The momentum value is 0.9. All these values were set empirically to optimize performance. We also found empirically that SVM works better with features from fc8 as opposed to fc6 and fc7.

\begin{table}
\centering

\begin{tabular}{| c | c | c | }
     \hline      
			Datasets									    &   Synthetic  Gradual    &  Synthetic Real      \\
     \hline
				 T2007t            & 19398	& 19439 	\\
		     T2007d            & 13656	& 19607		\\
				 T2008              & 39047	& 32456		\\
				 T2009              & 44158	& 32578 	\\ 
			\hline
			Total               & 116259 & 104080   \\ 
			\hline 
    \end{tabular}\vspace{3pt}
\caption{Our dataset SBD\_Syn in terms of number of segments (16 frames each). SBD\_Syn is synthetically generated from T2007t/d,2008,2009 using image compositing. The data contains a balanced portion of no-transitions.
}
\label{tab:DeepSBDSyn}
\end{table}


\begin{table}
\small
\centering
\begin{tabular}{| l | l | }
     \hline      
			Transitions									    &   Number of segments    \\
     \hline
				 No-transition              & 4,427 \\
			   Gradual              & 11,249 \\
				 Sharp            & 359 \\
			\hline
			Total               & 16,035 \\ 
			\hline
    \end{tabular}\vspace{3pt}
\caption{Our SBD\_BT dataset. The no-transition represent complex hard-negatives (Figure~\ref{fig:BTData}). When included in training, precision is significantly is improved.}
\label{tab:DeepSBDBT}
\end{table}

\section{Results}

We performed experiments on real data as well as on synthetically generated data. We examined 4,683,552 frames, $81.8\%$ of which are real. Our work is the largest SBD evaluation to date for one algorithm. We asses performance quantitatively using precision (P), recall (R) and F-score (F). Here, we use the standard TRECVID evaluation metric \cite{Smeaton10} where a transition is detected if it overlaps with the annotations by at least one frame. We report the per-transition performances. During comparison we highlight the best performing technique in \textbf{bold}. To account for possible mis-annotations and system error in such large experiment, we claim a technique is superior only if it achieves more then $0.5\%$ P, R, or F improvement over the second best performing technique. Techniques with $0.5\%$ difference are claimed as competitive. We train two models, both using our datasets DSB\_Syn and SBD\_BT. One of them uses few real data from T2005 and Barladi et al. , denoted by \textbf{$r$}, at most $7\%$ of the total training data. Both models are competitive to each other. We report results with \textbf{$r$} in the paper and report the other model in the supplementary material. 

We compare against the latest techniques (Lu et al. \cite{Lu13}, Priya et al. \cite{Priya14}, Apostolidis et al. \cite{Apostolidis14} and Berladi et al. \cite{Berladi15}) as well as the best performers in the 7 years of the TRECVID challenge (Yuan et al. \cite{Yuan04} and Liu et al. \cite{Liu07}). These techniques show the compromise between detection accuracy and processing speed commonly present in SBD. Lu et al. \cite{Lu13} is the fastest of all, but generates conservative performance. Priya et al. \cite{Priya14}, Liu et al. \cite{Liu07} and Yuan et al. \cite{Yuan04} generate better performance. However, at the cost of heavy optical flow computation. Our results show that DeepSBD optimizes both detection accuracy and processing speed over all current techniques. That is, we outperform gradual detection, generate competitive performance for sharp transitions and produce significant improvement in wipes detection. In addition, we are up to $11$ times faster than the state of the art. More detailed results are reported in the supplementary material. 

\subsection{Real Sequences} 

We evaluated our technique on all seven TRECVID releases, from 2001 to 2007. They have a total of 3,831,648 frames, with 8,545 gradual and 14,602 sharp transitions. No test data was included in the training. Table~\ref{tab:Ds} shows performance evaluation on 6 sequences commonly used in Lu et al. \cite{Lu13} and Priya et al. \cite{Priya14}. The sequences are from T2001a (see Table~\ref{tab:2001ab}) and present challenging videos from outer-space. The videos include cases such as global illumination variation, smoke, fire and fast non-rigid motion. We outperform Lu et al. in all sequences for both gradual and sharp transition. Furthermore, we outperform Priya et al. in the vast majority of sequences in both transition types. 

\begin{table}
\centering
\small
\begin{tabular}{|l | l |}
     \hline      
												                   & videos names       \\
			\hline 
				 T2001a                        & BOR10\_001, BOR10\_002, NAD57, NAD58, \\
				                               & anni001, anni005, anni006, anni007, anni00, \\
																			 & anni009, anni010   \\ 
						\hline															
				 T2001b                           & BOR03, BOR08, BOR10, BOR12, BOR17 \\ 
			\hline
    \end{tabular}\vspace{3pt}
\caption{Videos of T2001a and T2001b}
\label{tab:2001ab}
\end{table}

\begin{table*}
\small
\centering
\begin{tabular}{|l | l | l | l | l | l| l| }
     \hline      
												                          &   D1-anni5   & D2-anni6  &  D3-anni9    & D4-anni10      & D5-NAD57  & D6-NAD58     \\
     \hline
			 \textbf{Lu et al. \cite{Lu13}}             &        &        &        &         &          &         \\
					Abrupt                                  &   -    & 0.905  &  0.754 &  0.892  &   -      &  0.962 \\
			    Gradual                                 &   -    & 0.817  &  0.824 &  0.734  &   -      &  0.884 \\
			\hline
	    \textbf{Priya et al. \cite{Priya14}}           &        &        &        &         &          &         \\
			   Abrupt                                   &  \textbf{0.85}  & 0.911  &  0.842 &  0.897  &  0.945   &  \textbf{0.945} \\
				 Gradual                                  &  0.938 & 0.885  &  0.873 &  0.822  &  0.809   &  0.885 \\
			\hline 
			\textbf{DeepSBD (ours)}                            &        &        &        &         &          &   \\
				 Abrupt                                   &  0.818 & \textbf{0.988}  & \textbf{0.961}  &  \textbf{0.918}  &  \textbf{0.957}   &  0.904 \\
				 Gradual                                  &  \textbf{0.945} & 0.885  & \textbf{0.919}  &  \textbf{0.855}  &  \textbf{0.917}    &  \textbf{0.914} \\
			\hline
    \end{tabular}\vspace{3pt}
\caption{DeepSBD evaluation on 6 challenging sequences from TRECVID 2001 (D1-D6). Our technique outperforms Lu et al. and Priya et al. \cite{Priya14} in the vast majority of sequences. The improvement is more significant in gradual transitions.}
\label{tab:Ds}
\end{table*}

\begin{table*}
\small
\centering
\begin{tabular}{|l | l | l | l | l | l | l | l | l | l |}
     \hline      
			Size of test-data (in sequences)	& 9       &  10      &  11     & 12      & 13    & 14     & 15   & 16  & 17  \\
     \hline
			 \textbf{Priya et al. \cite{Priya14}}       &         &         &         &         &       &        &     &   &   \\
					Abrupt                                  &  0.9733 &  0.974 &  0.9748  &  0.9742 & 0.9737 & 0.9741  & 0.9733 & 0.9737 & 0.974\\
			    Gradual                                 &  0.775  &  0.7578  &  0.7677 &  0.7742 & 0.7811 & 0.7825 & 0.7802 & \textbf{0.7726} & \textbf{0.78} \\ 
			\hline
			\textbf{DeepSBD (ours)}                            &         &         &         &         &       &        &     &   &   \\
				 Abrupt                                   &  0.9729 &  0.9749  &  0.9743 &  0.974 & 0.9743 & 0.9749  & 0.9733   & 0.9713 & 0.9726\\ 
				 Gradual                                  &  \textbf{0.8962}  &  \textbf{0.8774} &  \textbf{0.8613} &  \textbf{0.8395}  & \textbf{0.8171} & \textbf{0.797}   & 0.7758 &  0.7507 & 0.7259\\ 
			\hline
    \end{tabular}\vspace{3pt}
\caption{Per-sequence f-score on T2007. We compare against Priya et al. \cite{Priya14} on test-sets of different sizes. Since Priya et al. \cite{Priya14} is trained on 7 out of the total 17 test sequences, comparison should be done on at most 10 sequences. Results show we significantly outperform Priya et al. with a test-set of 10 sequences. Here, our gradual transitions detector is more than $12\%$ better than Priya et al. in f-score, a significant improvement due to our CNN solution. Furthermore, we still outperform Priya et al. with a test-set size up to 14 sequences. Here, however, at least 4 sequences were included in Priya et al. which biases the results towards Priya et al. Including these videos in our training is expected to boost our performance even further.}
\label{tab:PriyaSummary}
\end{table*}

Table~\ref{tab:PriyaSummary} compares our technique against Priya et al. \cite{Priya14} on T2007. Note that \cite{Priya14} used a slightly  different approach for evaluation than the one recommended by TRECVID \cite{Smeaton10}. TRECVID recommends estimating the average performance per transition. However, \cite{Priya14} estimated the average performance per sequence. Furthermore, Priya et al. tested on 17 sequences, 7 of which were included in their training set. This biases the results towards Priya et al. \cite{Priya14}. Hence, for fair comparison these 7 sequences should be removed from the 17 test sequences and the comparison should be done on at most 10 sequences. To illustrate this point, we examined our technique with different sizes of the test dataset. Each column of Table~\ref{tab:PriyaSummary} shows the performance with different size of the test data. With 10 test sequences, our technique outperforms Priya et al. [16] significantly in gradual transitions (0.88 vs. 0.76 f-score) and generates competitive results for sharp transitions. Furthermore, we still outperform Priya et al. even with a test-set of 14 sequences. Here, however, at least 4 sequences are included in Priya et al. training and hence results are biased towards Priya et al. Including these videos in our training is expected to improve performance even further. The spatio-temporal aspect of our solution allow us to generate these high detection accuracy results without explicitly estimating optical flow. Our experimental results showed that just relying on the spatial information generates very poor performance.

\begin{figure*}
  \centering
   \includegraphics[height=4cm, width=18cm]{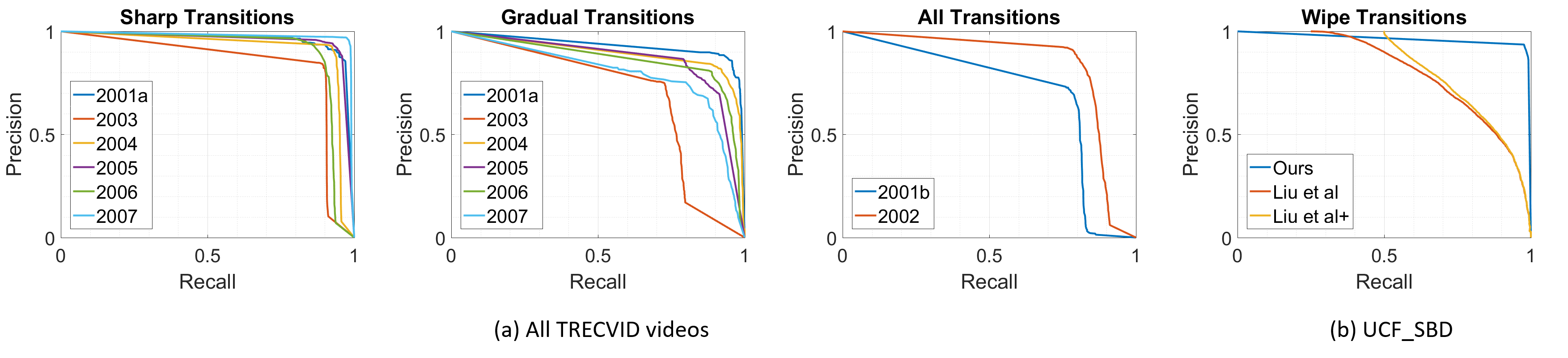}
   \caption{(a) Precision-Recall of DeepSBD for all TRECVID sequences. (b) Precision-Recall of DeepSBD for wipes. Here, we compare against two implementations of Liu et al. \cite{Liu07} wipe detector. The first implementation examines all frames of UCF101\_SBD and `+' examines only frames not classified as sharp nor gradual by DeepSBD. We significantly outperforms both approaches.}
\label{fig:AllROC}
 \end{figure*}

Table~\ref{tab:TRECSVD} evaluates DeepSBD on T2004, 2005, 2006 and 2007. To test on 2005, we removed it from our training. We compare against the best TRECVID performers as well as Lu et al. \cite{Lu13}. We significantly outperform Lu et al. in T2007. Furthermore, we outperform the best TRECVID performers, Liu et al. \cite{Liu07} and Yuan et al. \cite{Yuan04} on all four datasets.
Table~\ref{tab:TRECRemaining} evaluates DeepSBD on the remaining TRECVID datasets. T2001b and 2002 annotations contain significant overlap between sharp and gradual transitions. Hence, for them we show the overall combined transitions performance. Furthermore, T2003 is missing 4 videos and hence we could not compare against the reported TRECVID performance. In all sequences we generate good performance. T2001b and 2002 sequences contain strong noise and jitter. Yet, our technique was robust enough to handle such artifacts. Figure~\ref{fig:AllROC} (a) shows the precision-recall curves for our DeepSBD on all real TRECVID sequences. Table~\ref{tab:RAI} shows the combined f-score for the RAI dataset \cite{RAIData}. Here, we compare against the techniques of Apostolidis et al. \cite{Apostolidis14} and Berladi et al. \cite{Berladi15}. Results show that we significantly outperform both techniques. The supplementary material (Table~II-XVI) shows the per sequence results for each of the TRECVID and RAI dataset examined by our technique. This includes much more statistics e.g. true positives (TP), false positives (FP), false negative (FN) and so on. 

\begin{table}
\small
\centering
\begin{tabular}{|l | l | l | l | l |}
     \hline      
												                   & T2004      & T2005 & T2006   &  T2007  \\
     \hline
			 \textbf{Best TRECVID}   &        &        &      &        \\
			\textbf{performers \cite{Smeaton10}}    &        &        &    &    \\
					Abrupt                           & 0.929   & 0.935 & 0.899  & 0.972  \\
			    Gradual                          & 0.806   & 0.786 & 0.814  & 0.753  \\ 
			\hline
	    \textbf{Lu et al. \cite{Lu13}}       &      &     &     &        \\
			   Abrupt                            &  -   &  -   & -   & 0.761  \\
				 Gradual                           &  -   &  -   & -   & 0.618  \\ 
			\hline 
			\textbf{DeepSBD (ours)}                     &            &       & &        \\
				 Abrupt                            &  0.926     & 0.934 & 0.895 & 0.971  \\ 
				 Gradual                           & \textbf{0.866}     & \textbf{0.844} & \textbf{0.827}  & \textbf{0.776}  \\ 
			\hline
    \end{tabular}\vspace{3pt}
\caption{Comparing DeepSBD against different techniques. We compare against TRECVID best performers, Liu et al. \cite{Liu07} for 2006/2007 and against Yuan et al. \cite{Yuan05} for 2004/2005. We also compare against the latest no optical flow technique of Lu et al. \cite{Lu13}. We outperform all techniques on all datasets.}
\label{tab:TRECSVD}
\end{table}

\begin{table}
\small
\centering
\begin{tabular}{|l | l | l | l |l|}
     \hline      
												                   & T2001a      & T2001b   &  T2002  & T2003 \\
			\hline 
			\textbf{DeepSBD}                     &            &        &       &  \\
				 Abrupt                            & 0.931     & -  & -  & 0.866\\ 
				 Gradual                           & 0.904     & -  & -  & 0.759\\ 
				 Overall                           & 0.918     &   0.748     &  0.865  & 0.8337 \\
			\hline
    \end{tabular}\vspace{3pt}
\caption{Evaluating DeepSBD on T2001a, 2001b, 2002 and 2003.}
\label{tab:TRECRemaining}
\end{table}


%

\begin{table}
\small
\centering
\begin{tabular}{|l | l | l | l |}
     \hline      
												                   & Method in \cite{Apostolidis14}      & Method in \cite{Berladi15}   &  Ours  \\			
			\hline
			 RAI \cite{RAIData}                         & 0.84     & 0.84       &  \textbf{0.94}  \\
			\hline
    \end{tabular}\vspace{3pt}
\caption{Processing the RAI dataset \cite{RAIData} with different techniques. Here we show the combined f-score of the overall detection. Our technique significantly outperforms both Apostolidis et al. \cite{Apostolidis14} and Berladi et al. \cite{Berladi15}.}
\label{tab:RAI}
\end{table}

\begin{figure}
  \centering
   \includegraphics[height=3.5cm, width=8cm]{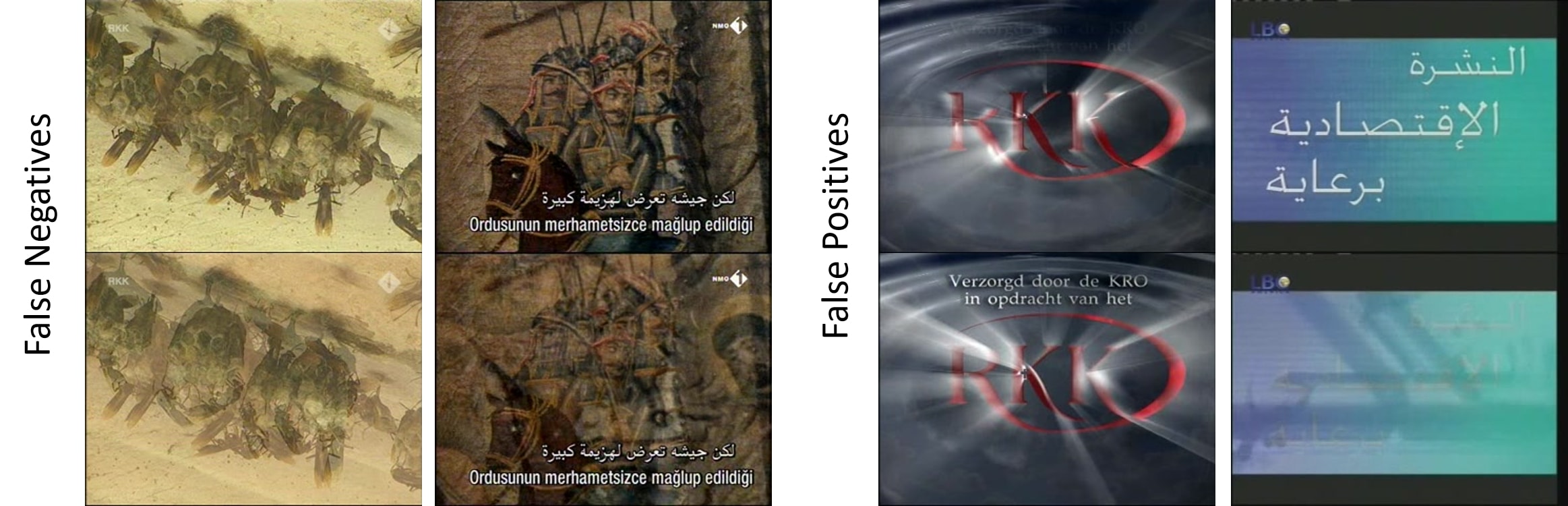}
   \caption{Failure cases in gradual transition detection.}
\label{fig:FailureG}
 \end{figure}

Table~\ref{tab:DifferentParameters} examines different test configurations for DeepSBD. SVM on fc8 generates better results than on fc6. The post-processing (pp) step improves the performance, especially for T2007. The best performance is obtained with fc8+svm+pp. Figure~\ref{fig:FailureG} shows failure cases in gradual transition detection. Too long transitions can get misclassified as False negatives (FN). Here, no enough temporal difference is captured over our 16 frames window. FN can also be generated when both shots have similar texture and color. False positives are largely generated by computer graphics content. Such content have a gradual-like effect. However, they are not semantically classified as a shot transition.  

\begin{table}
\small
\centering
\begin{tabular}{| l | l | l | l | l | l | }
     \hline              
												    &   fc6+svm   & fc8+svm  &  fc8+svm+pp & fc8+pp      \\
     \hline
		\textbf{TR2001a} & & & &\\ 
		
	      Gradual   		          & 0.882  & \textbf{0.906}	& \textbf{0.904}	& \textbf{0.906} 	 \\
				
				Sharp                 & \textbf{0.931}	& \textbf{0.931}	& \textbf{0.931}	& 0.923	 \\
				\hline 		
		\textbf{TR2006} & & & &\\ 
		
	      Gradual   		          & 0.83  & \textbf{0.841}	& \textbf{0.844}	& \textbf{0.843} 	 \\
				
				Sharp                 & 0.887	& \textbf{0.895}	& \textbf{0.895}	& \textbf{0.895}	 \\
				\hline 		
						\textbf{TR2007} & & & &\\ 
		
	      Gradual   		          & 0.71  & 0.732	& \textbf{0.776}	& \textbf{0.776} 	 \\
				
				Sharp                 & 0.955	& 0.968	& \textbf{0.971}	& \textbf{0.973}	 \\
				\hline 	
							
    \end{tabular}\vspace{3pt}
\caption{Comparing DeepSBD with different settings. The best performance on all all datasets is obtained with fc8 features + svm + post-processing (pp). Some other settings are competitive (see bold).}
\label{tab:DifferentParameters}
\end{table}

\subsection{The importance of our datasets}

Table~\ref{tab:OurDataImpact} shows the significance and importance of our datasets SBD\_Syn and SBD\_BT in generating high accuracy detections. We evaluate DeepSBD on T2007 with six different training sets: 1) R\_3-5 2) R\_3-6 3) R\_3-6 + BT, 4) S + r, 5) S + r + BT and 6) and S + BT. S and BT is short for our datasets SBD\_Syn and SBD\_BT. R\_3-6 represent TRECVID real videos and annotations from 2003 to 2006. $r$ is T2005 and Baraldi \cite{Baraldi15}. Results show that training with R\_3-5  generate poor performance. In addition, it limits us to testing on just 3 data-sets (T2001a, T2006 and T2007). Adding T2006 to training improves performance but limits our testing further to 2 data-sets (T2001a and T2007). Adding our bootstrapping data SBD\_BT (BT) improves precision and performance significantly. This shows the high quality and importance of our SBD\_BT. The best performance, however, is generated when both our datasets SBD\_Syn and SBD\_BT with $r$ are used for training. In addition to the highest performance, this option allow us to test on all TRECVID videos, except T2005. Removing $r$ from the training generates the second best performance. This, however, allow us to test on all TRECVID videos, including T2005. The experiment shows the significance and importance of our data-sets. We performed this experiment on several test sets and we found S + r + BT and S + BT are always the top and competitive to each other (see supplementary material, Table.~1). This shows the significance of our datasets and their generation process (Section~\ref{Sec:Dataset}).  

\begin{table}
\small
\centering
\begin{tabular}{| l | l | l | l | l | l | l| l| }
     \hline              
												    &   P   & R  &  F & P    & R      & F       \\
     \hline
				%

		   \hline
	      R\_3-5   		   & 0.495  & 0.665	& 0.568	& 0.894 	& 0.872	& 	0.883 \\
				\hline
				R\_3-6         & 0.683	& 0.683	& 0.683	& 0.957	& 0.95	& 0.953	 \\
				 \hline
				R\_3-6 + BT   & 0.755 & 0.705 & 0.729 & 0.961 & 0.961 & 0.961 \\
				\hline
				S + r         & 0.722 & 0.63 & 0.673 & 0.979 & 0.955 & \textbf{0.967} \\
        \hline
				S + r + BT    & \textbf{0.799} & \textbf{0.753} & \textbf{0.776} & \textbf{0.973} & \textbf{0.969} & \textbf{0.971}\\				
        \hline
			  S + BT     & 0.779 & 0.714 & 0.745 & 0.969 & 0.966 & \textbf{0.968}\\	
				\hline
    \end{tabular}\vspace{3pt}
\caption{Training DeepSBD with different datasets. Results show that best performance is generated when our both SBD\_Syn (S) and SBD\_BT (BT) and r are used. Removing any real sequences ($r$) from our datasets generates the second best performance (S+BT). The advantage of this option is allowing us to test on all TRECVID videos, including T2005 (Table~\ref{tab:TRECSVD}). The table also shows SBD\_BT clearly improves the precision and overall performance.}
\label{tab:OurDataImpact}
\end{table}

\subsection{Controlled Experiments} 

\begin{figure}
  \centering
   \includegraphics[height=3cm, width=7cm]{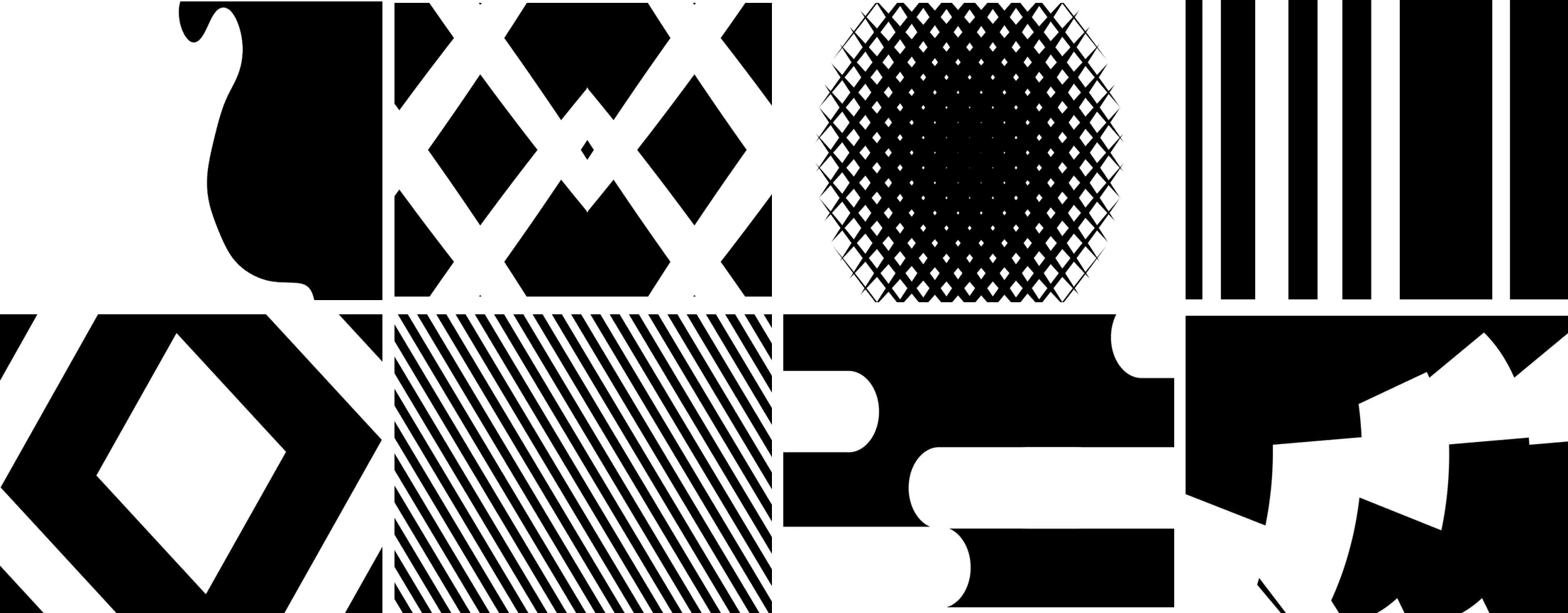}
   \caption{Samples of the $\alpha$ mattes used for generating the wipes of UCF101\_SBD. Examples of the generated wipes are shown in the supplementary material in Figure 3.}
\label{fig:Wipes}
 \end{figure}

We generated a synthetic test-set. Our dataset contain 53,324 segments, divided equally between gradual, sharp, wipes and no-transitions. Each segment is 16 frames long. We generated gradual and sharp transition using image compositing as we did for SDB\_Syn (see Eq.~\ref{eq:DataGen}). Here, we constrain the shots to come from two different UCF101 videos \cite{Soomro12}. We present the first large wipes dataset, containing 1.1 million wipe frames ($20\%$ test). They are also generated using Eq.~\ref{eq:DataGen}. Here, however, the opacity values $\alpha$ have more complicated spatio-temporal patterns than sharp and gradual transitions. Figure~\ref{fig:Wipes} shows some of the 196 $\alpha$ mattes we used. The supplementary material, Figure~3, show frames from our wipes dataset. We call our synthetic UCF dataset UCF101\_SBD.      

We train the model using SBD\_Syn, SBD\_BT and the synthetic wipes. This model generates 4 classes. Table~\ref{tab:UCF101_SBD} evaluates DeepSBD on UCF101\_SBD. 
We generate high performance for all classes, including wipes. Performance is higher than the ones previously reported on the TRECVID sequences. This could be due to the highly accurate annotations of UCF101\_SBD. Figure~\ref{fig:AllROC} (b) compares our wipe detector against the state of the art of Liu et al. \cite{Liu07}. We evaluate Liu et al. using two strategies. The first examines all frames of UCF101\_SBD. The second, `+', examines only frames not detected as gradual nor sharp transitions by DeepSBD. Our technique outperform both approaches significantly.  

\begin{table}
\small
\centering
\begin{tabular}{| l | l | l | l | l | }
     \hline      
			Transition Type							    & Precision      & Recall   & F-measure     \\
     \hline
					Gradual              	& 0.989	& 0.995	& 0.992 \\
			   Sharp                 & 0.984	& 0.999	& 0.992 \\
				 Wipes                & 0.976	& 0.936	& 0.956 \\
			\hline 
    \end{tabular}\vspace{3pt}
\caption{Evaluating DeepSBD on our synthetic dataset UCF101\_SBD. In total UCF101\_SBD has $53,253$ segments, divided equally among all classes (normal, gradual, sharp and wipes). Each segment has 16 frames. DeepSBD generates a very high performance in all classes.}
\label{tab:UCF101_SBD}
\end{table}

%

\subsection{Processing Speed}
%

\begin{table}
\small
\centering
\begin{tabular}{|l | l |}
     \hline      
												                & Real-time speed-up factor \\
			\hline
			DeepSBD                           & 19.3 \\
      \hline 
			Liu et al. \cite{Liu07,Smeaton10} & 3.24    \\
      \hline
		  Priya et al. \cite{Priya14}   & 1.76  \\
			\hline
			Yuan et al. \cite{Yuan05} & 2.43 \\
			\hline
    \end{tabular}\vspace{3pt}
\caption{Real-time speed-up factor for different shot boundary detection techniques. We are faster than all techniques with a factor up to 11.}   
\label{tab:ProcessingTime}
\end{table}

We examined a TRECVID video of duration 4,096 seconds containing 102,400 frames. We ran the test-phase of DeepSBD with different batch sizes as input. The GPU performs n iterations until all 102,400 frames are processed. The smaller the batch size, the more iterations required and hence the more time required to process all frames. However, the less memory required. Experiments shows that the processing speed gain from 10 to 100 batch size is not significant. That is between 16-19.3 real-time speed up factor. 
We use Titan X, a GPU commonly used for deep learning applications. Table~\ref{tab:ProcessingTime} compares the processing speed of different SBD techniques. In comparison with the best performing optical-flow based techniques, we are 11 times faster than Priya et al. \cite{Priya14}, 6 times faster than Liu et al. \cite{Liu07} and 9.65 times faster than Yuan et al. \cite{Yuan05}. The supplementary material shows more analysis of the processing speed in Figure~4-5 (Section II). 

%

\subsection{Deep Analysis on Network Responses} 
\begin{figure}
  \centering
   \includegraphics[height=22cm, width=9cm]{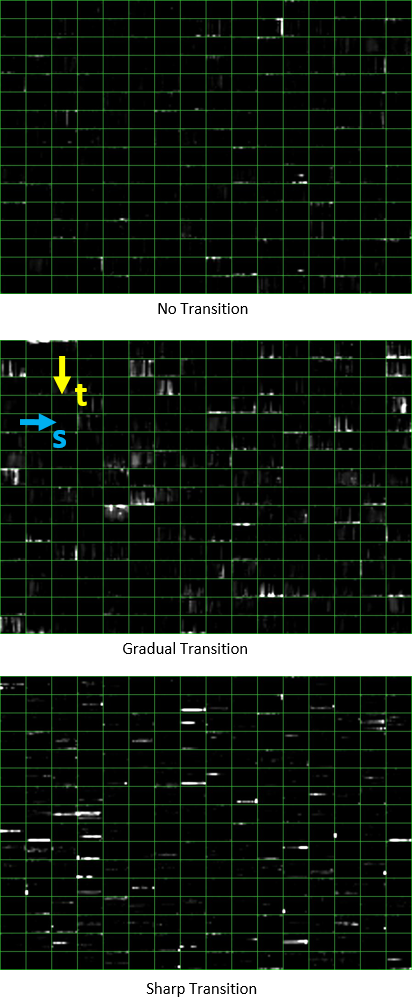}
   \caption{Filter responses of DeepSBD stacked next to each other. The green grid shows filters' borders. Here, y-axis is time (see blue arrow) and x-axis is space (see yellow arrow). Sharp transitions have an abrupt response in time (bright horizontal lines). Gradual transitions have blurred responses in time. No transition do not show specific patterns. More examples are shown in the supplementary material in Figure~6.}
\label{fig:FilterResponses}
 \end{figure}

We randomly selected two segments (16 frames) from UCF101 and synthetically generated a sharp and gradual transition using Eq.~\ref{eq:DataGen}. We treated one of the two sequences as no-transition. We examined all segments using DeepSBD. 
Figure~\ref{fig:FilterResponses} shows the heat map of some Conv5 filter responses for each transition type. The filters are stacked next to each other, in blocks. The green grid shows filters' borders. Time is the y-axis and space is the x-axis. Vertical space is averaged over the horizontal space. Sharp transitions have abrupt responses in the time axis in form of bright horizontal lines. Gradual transitions have blurred responses in the time axis. No transitions do not show a specific response pattern. The patterns are consistent on several other segments. The supplementary material shows more of such results in Figure~6 (Section III). 



\section{Conclusion}

We presented the first CNN technique for shot boundary detection. Current techniques compromise between detection accuracy and processing speed and use hand-crafted features. We exploit big data to optimize both accuracy and speed. This is important as SBD is a common pre-processing step for video manipulation. We present two large datasets containing 3.57 million frames. One set is generated synthetically while the other is carefully annotated through bootstrapping. We outperform state of the art gradual transition detections, generate competitive performance in sharp transitions and produce significant improvement in wipes detections. Our approach is up to 11 times faster than the state of the art. Future work can examine computer graphics content more closely. We will release our datasets and code to encourage future research.



\ifCLASSOPTIONcaptionsoff
  \newpage
\fi



%




{
\bibliographystyle{IEEEtran}
\bibliography{references}
}

%





\newpage
\title{Supplementary Material: Large-scale, Fast and Accurate Shot Boundary Detection through Spatio-temporal Convolutional Neural Networks}
\maketitle
\setcounter{equation}{0}
\setcounter{figure}{0}
\setcounter{table}{0}
\setcounter{section}{0}
\setcounter{page}{1}
\makeatletter
\renewcommand{\theequation}{\arabic{equation}}
\renewcommand{\thefigure}{\arabic{figure}}

\section{Our data-sets}

\begin{table}[h]
\small
\centering
\begin{tabular}{| l | l | l | l | l | l | l| l| }
     \hline              
   &   P   & R  &  F & P    & R      & F       \\
     \hline
\textbf{T2001a} & & & & & & \\ 
\hline
     R\_3-5  
         & 0.693  & 0.78
& 0.734  & 0.863 & 0.691
& 0.768 \\
\hline
R\_3-6                & 0.762
& 0.814  & 0.787 & 0.93
& 0.891  & 0.91 \\
\hline
R\_3-6+BT  & 0.917 & 0.753 & 0.827 & 0.96 & 0.923 & \textbf{0.941} \\
\hline
S+r & 0.782 & 0.851 & 0.815 & 0.926 & 0.92 & 0.923 \\
\hline
S+r+BT  & \textbf{0.951} & 0.861 & 0.904 & 0.927 & \textbf{0.936} & 0.931\\
\hline
S+BT  & 0.934 & \textbf{0.912} & \textbf{0.923} & \textbf{0.979} & 0.904 & \textbf{0.94}\\
\hline 

\textbf{T2006} & & & & & & \\ 
\hline
     R\_3-5    
         & 0.641  & 0.747
& 0.69  & 0.691 & 0.838
& 0.758 \\
\hline
S+r & 0.834 & 0.744 & 0.786 & 0.86 & 0.873 & 0.866 \\
\hline
S+r+BT   & \textbf{0.888} & 0.804 & \textbf{0.844} & 0.863 & \textbf{0.93} & 
\textbf{0.895}\\
\hline
S+BT  & 0.827 & \textbf{0.834} & 0.83 & \textbf{0.876} & 0.869 & 0.872\\
\hline

 \textbf{T2007} & & & & & & \\ 
\hline
     R\_3-5            & 0.495  & 0.665
& 0.568  & 0.894 & 0.872
& 0.883 \\
\hline
R\_3-6 + & 0.683
& 0.683  & 0.683 & 0.957
& 0.95  & 0.953 \\
\hline
R\_3-6+BT & 0.755 & 0.705 & 0.729 & 0.961 & 0.961 & 0.961 \\
\hline
S+r & 0.722 & 0.63 & 0.673 & 0.979 & 0.955 & \textbf{0.967} \\
\hline
S+r+BT  & \textbf{0.799} & \textbf{0.753} & \textbf{0.776} & \textbf{0.973} & \textbf{0.969} & \textbf{0.971}\\
\hline
S+BT  & 0.779 & 0.714 & 0.745 & 0.969 & \textbf{0.966} & \textbf{0.968}\\
\hline
\textbf{T2003} & & & & & & \\ 
\hline
S+r & 0.735 &	0.703 &	0.718 &	\textbf{0.899} &	0.837 &	0.867\\
\hline
S+r+BT & \textbf{0.779} &	0.741 &	0.759 &	0.892 &	\textbf{0.842} &	0.866\\
\hline
S+BT  & 0.741 &	\textbf{0.804} &	\textbf{0.771} &	\textbf{0.898} &	\textbf{0.846} &	\textbf{0.871}\\
\hline
\textbf{T2004} & & & & & & \\ 
\hline
S+r & 0.868	& 0.774 &	0.818	&	0.928 &	\textbf{0.929} &	\textbf{0.929} \\
\hline
S+r+BT & \textbf{0.918} &	0.819 &	0.866 &	0.923 &	\textbf{0.929} &	\textbf{0.926}\\
\hline
S+BT  & 0.888 &	\textbf{0.884} &	\textbf{0.886} &	\textbf{0.941} &	0.918 &	\textbf{0.929}\\
\hline
\textbf{T2005} & & & & & & \\ 
\hline
S+BT & 0.791 &	0.866 &	0.827 &	0.927 &	0.941 &	0.934 \\
\hline

\end{tabular}
\caption{Training our technique DeepSBD with different datasets. R\_3-5 represent all TRECVID videos except 2001a, 2006 and 2007.  Results show that the best performance is always generated when both our synthetic (S) and bootstrapping (BT) datasets are used (see S+r+BT and S+BT). Here, r is a very small portion of real videos (T2005 and Baraldi). The advantage of using S+BT is allowing us to test on all TRECVID videos, including T2005. Finally, our bootstrapping data BT clearly improves the precision and overall performance.}
\label{tab:OurDataImpact}
\end{table}

\begin{figure*}[!t]

\includegraphics[scale=.15]{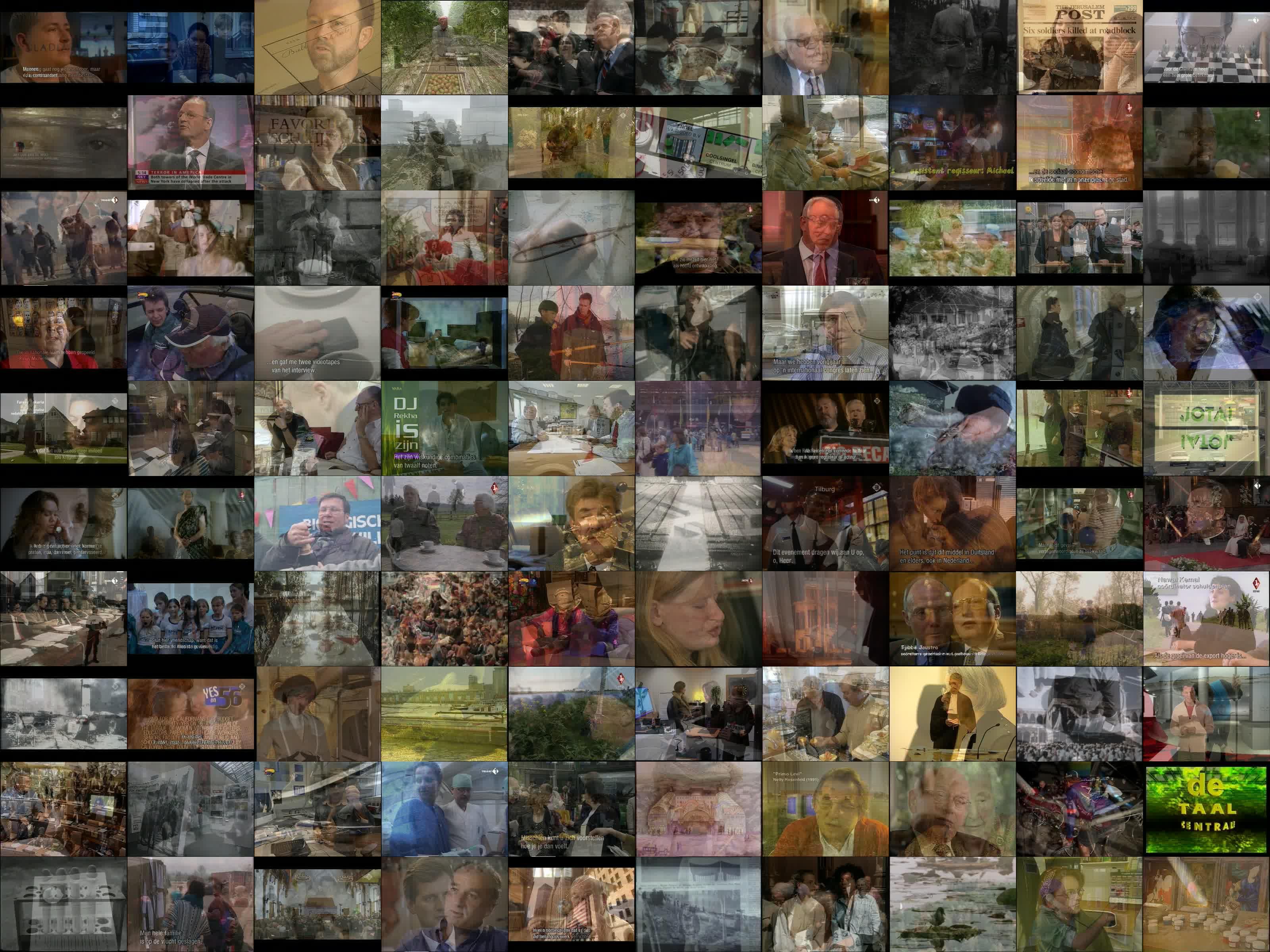}

\centering
\caption{100 frames from the gradual transition class of our dataset. This data is generated synthetically through image compositing.}
\label{Mosaic_gradual}

\end{figure*}

\begin{figure*}[!t]

\includegraphics[scale=.15]{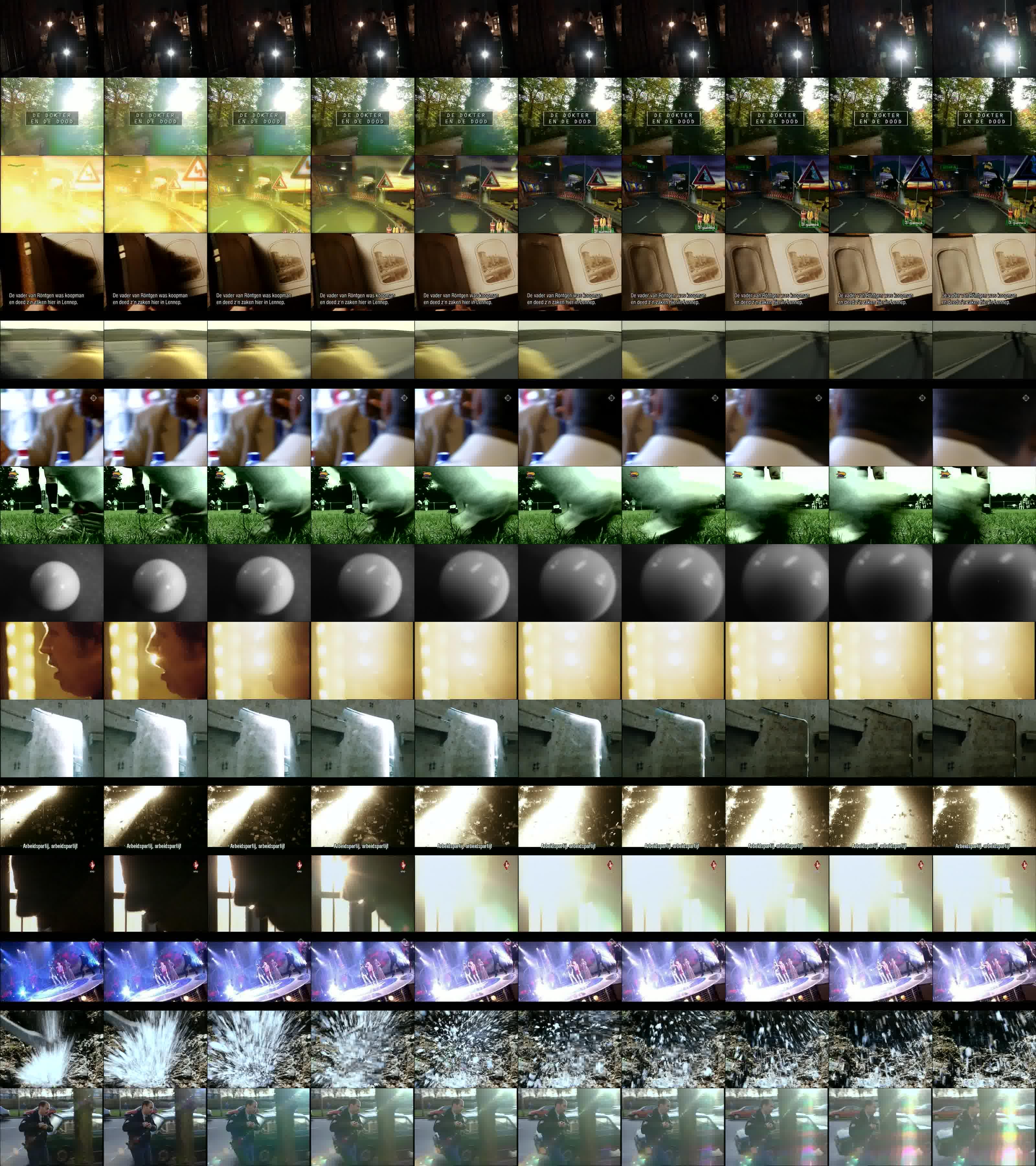}

\centering
\caption{Hard negative samples from our bootstrapping dataset. We carefully selected these samples through a semi-automated process. They represent complicated cases such as illumination variation, fast motion, occlusion and so on.}
\label{Mosaic_BT}

\end{figure*}

\begin{figure*}[t]

\includegraphics[scale=.15]{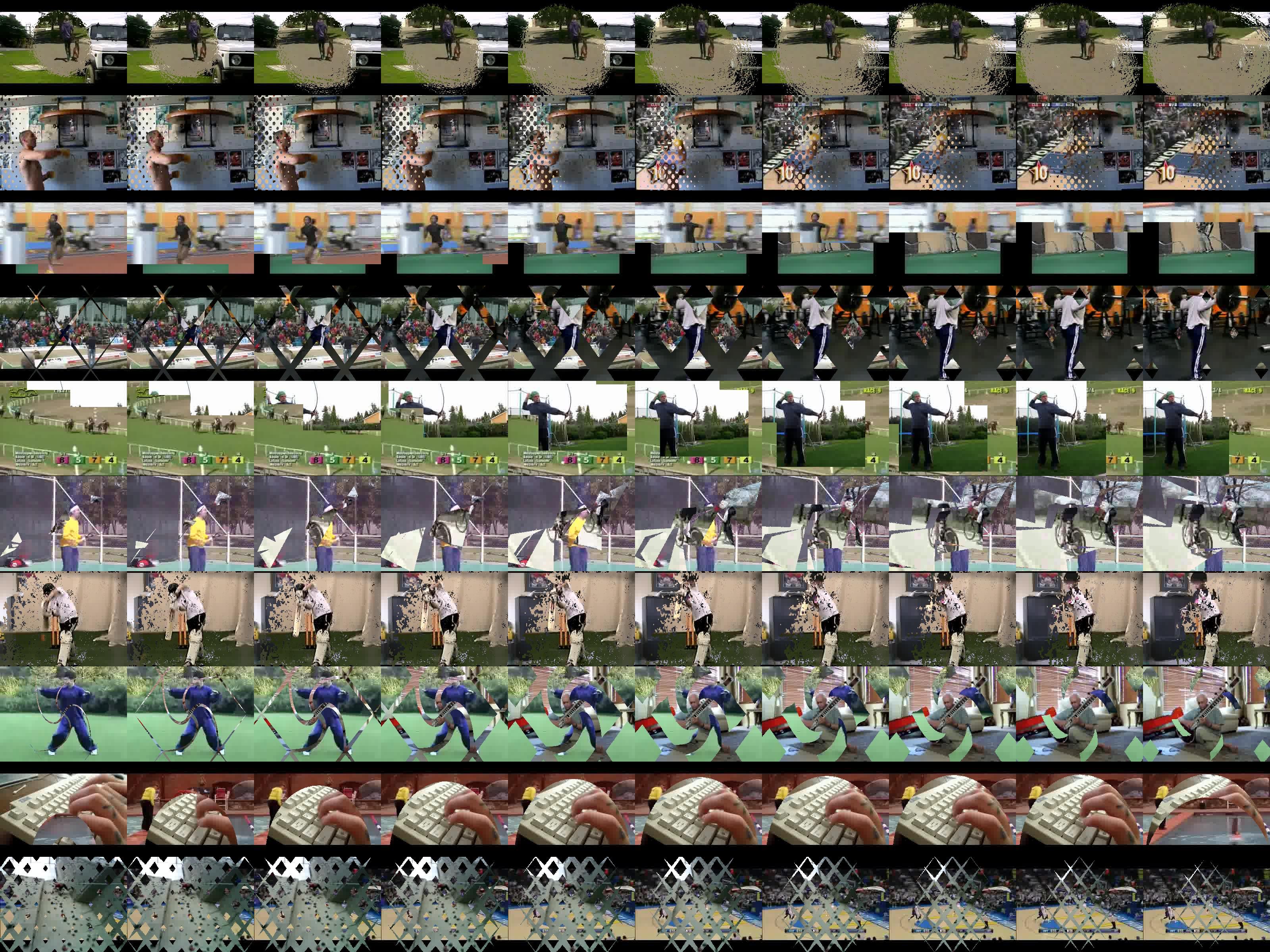}

\centering

\caption{10 sequences from our synthetically generated wipes data-set. Each row shows frames from one sequence. The dataset is generated using image compositing models with a wide variety of alpha mattes.}
\label{mosaic_wipe}
\end{figure*}

Figure~\ref{Mosaic_gradual} shows samples from the gradual transitions class of our dataset (SBD\_Syn). Our data is synthetically generated through image compositing. It is diverse, containing a wide variety of colors, texture, objects, motion and so on. Figure~\ref{Mosaic_BT} shows hard negative samples from our bootstrapping data (SBD\_BT). The samples contain challenging cases that commonly confuse gradual transition detectors e.g. fast motion, fast zoom in, illumination changes, object occlusion, strong lighting, and so on. Figure~\ref{mosaic_wipe} shows 10 sequences from our synthetically generated wipes dataset. The sequences show some variety of the alpha mattes used to generate wipes. 
 


Tab.~\ref{tab:OurDataImpact} shows the significance and importance of our synthetic SBD\_Syn and bootstrapping SBD\_BT datasets in generating high accuracy detections. We evaluate our technique, DeepSBD, on different datasets with six different training sets: 1) R\_3-5 2) R\_3-6 3) R\_3-6 + BT, 4) S + r, 5) S + r + BT and 6) and S + BT. S and BT is short for our datasets SBD\_Syn and SBD\_BT. R\_3-6 represent TRECVID real videos and annotations from 2003 to 2006. $r$ is T2005 and Baraldi. Results show that training with R\_3-5  generate poor performance. In addition, it limits us to testing on just 3 data-sets (T2001a, T2006 and T2007). Adding T2006 to training improves performance but limits our testing further to 2 data-sets (T2001a and T2007). Adding our bootstrapping data SBD\_BT (BT) improves precision and performance significantly. This shows the high quality and importance of our SBD\_BT. The best performance, however, is generated when both our datasets SBD\_Syn and SBD\_BT with $r$ are used for training. In addition to the highest performance, this option allow us to test on all TRECVID videos, except T2005. Removing $r$ from the training generates a competitive performance. This, however, allow us to test on all TRECVID videos, including T2005. The experiment shows the significance and importance of our data-sets. We performed this experiment on several test sets and we found S + r + BT and S + BT are always the top and competitive to each other. This shows the significance of our datasets. 

Tab.~\ref{2001b_07}-\ref{tab:RAI} shows detailed per video results for different testing sets. For each testing dataset, we report the results using two different training-sets (S+r+BT and S+BT). We show: the number of transitions (\#T), true positives (TP), false positives (FP), false negatives (FN), precision (P), recall (R) and F-measure (F).

\section{Processing speed}

\begin{figure}[t]

\includegraphics[scale=.5]{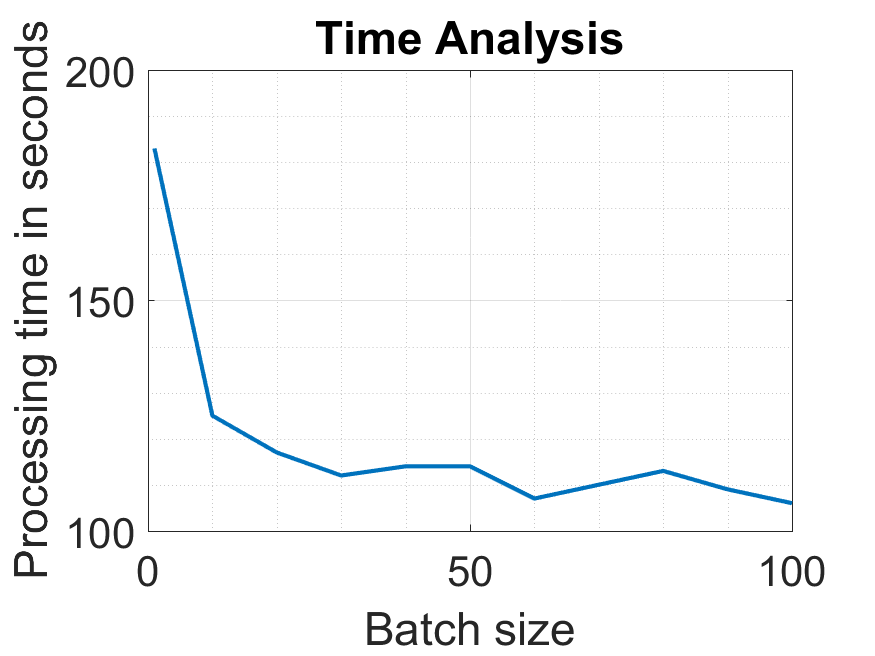}

\centering

\caption{Processing time in seconds of our technique. We report the results for different batch sizes as input.}
\label{PT_secs}
\end{figure}

\begin{figure}[t]

\includegraphics[scale=.5]{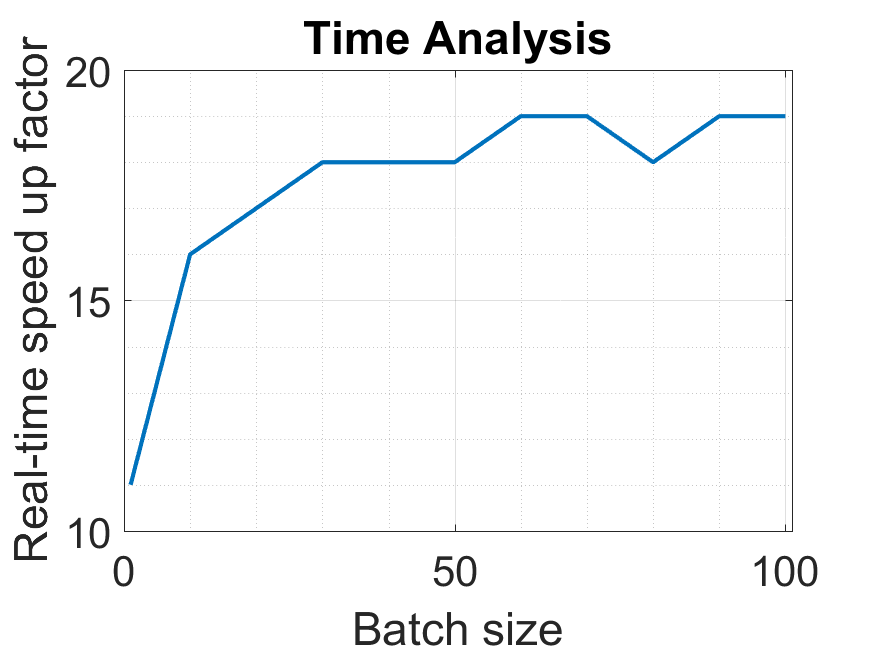}

\centering

\caption{Real-time speed factor of our technique. We report the results for different batch sizes as input.}
\label{SUF}
\end{figure}

Figure~\ref{PT_secs}-\ref{SUF} examines the processing speed (test-phase) of our technique with different batch sizes as input. We ran our model on 6,394 segments. Each segment is 16 frames long, and hence our test-set contains 102,304 frames. Figure~\ref{PT_secs} reports the total processing speed in seconds while Figure~\ref{SUF} reports the real-time speed up factor. Tab.~\ref{tab:PT} shows detailed analysis of this experiment. For each batch size we ran our technique twice to ensure consistency. Results show that the processing speed gain from 10 to 100 batch size is not significant. That’s between 16-19.3 real-time speed up factor. 


%

\section{Deep Analysis on Network Responses}

\begin{figure*}[t]
  \centering  
   \includegraphics[height=22cm, width=18cm]{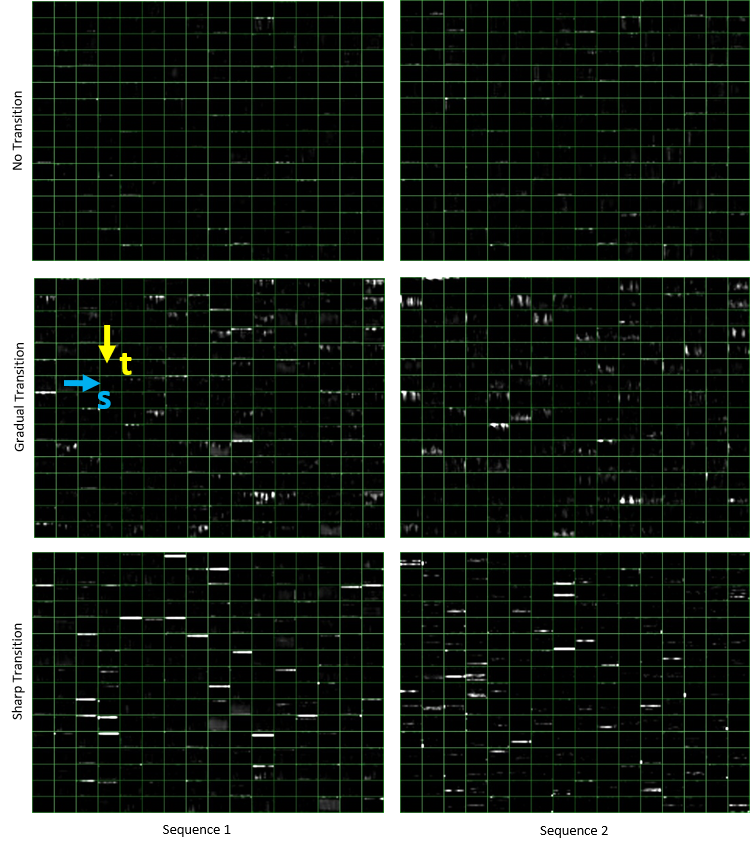}
  \caption{Filter responses of our technique DeepSBD stacked next to each other. The green grid shows filters' borders. Here, y-axis is time (see blue arrow) and x-axis is space (see yellow arrow). Sharp transitions have an abrupt response in time (bright horizontal lines). Gradual transitions have blurred responses in time. No transition do not show specific patterns.}
  \label{fig:FR}
\end{figure*}

Figure~\ref{fig:FR} visualizes the feature response of our technique. We show the visualization of two different image sequences. For each sequence, we randomly selected two segments (16 frames) from UCF101 and synthetically generated a sharp and gradual transition using image compositing models. We treated one of the two sequences as no-transition. We examined all segments using our technique, DeepSBD. 
Figure~\ref{fig:FR} shows the heat map of some Conv5 filter responses for each transition type. The filters are stacked next to each other, in blocks. The green grid shows some filters' borders. Time is the y-axis and space is the x-axis. Vertical space is averaged over the horizontal space. Sharp transitions have abrupt responses in the time axis in form of bright horizontal lines. Gradual transitions have blurred responses in the time axis. No transitions do not show a specific response pattern. The learned patterns of the three classes capture meaningful and discriminative information for the different types of shot transitions. Such information generate high detection accuracy as shown through out our results.



\begin{table}[h]
\centering
\begin{tabular}{|c|c|c|c|c|c|c|c|}
\hline
& \multicolumn{6}{c|}{\textbf{Gradual and Sharp}}  \\ \hline
\textbf{Video} & \textbf{TP} & \textbf{FP} & \textbf{FN} & \textbf{P} & \textbf{R} & \textbf{F} \\ \hline
\textbf{BOR03}     & 237         & 32          & 5           & 0.881      & 0.979      & 0.928      \\ \hline
\textbf{BOR08}     & 456         & 8           & 75          & 0.983      & 0.859      & 0.917      \\ \hline
\textbf{BOR10}     & 58          & 84          & 94          & 0.408      & 0.382      & 0.395      \\ \hline
\textbf{BOR12}     & 117         & 5           & 19          & 0.959      & 0.86       & 0.907      \\ \hline
\textbf{BOR17}     & 77          & 137         & 171         & 0.36       & 0.31       & 0.333      \\ \hline
\textbf{Total}     & 945         & 266         & 364         & 0.78       & 0.722      & 0.75       \\ \hline
\end{tabular}

\caption{Detailed per video results of T2001b. Here, we use S+r+BT for training our model. We report the combined results for both gradual and sharp transitions. We show the true positives (TP), false positives (FP), false negatives (FN), precision (P), recall (R) and F-measure (F).}
\label{2001b_07}
\end{table}

\begin{table}[h]
\centering

\begin{tabular}{|c|c|c|c|c|c|c|c|}
\hline
& \multicolumn{6}{c|}{\textbf{Gradual and Sharp}}  \\ \hline
\textbf{Video}  & \textbf{TP} & \textbf{FP} & \textbf{FN} & \textbf{P} & \textbf{R} & \textbf{F} \\ \hline
\textbf{BOR03}   & 240         & 30          & 2           & 0.889      & 0.992      & 0.938      \\ \hline
\textbf{BOR08}   & 500         & 7           & 31          & 0.986      & 0.942      & 0.963      \\ \hline
\textbf{BOR10}   & 54          & 82          & 98          & 0.397      & 0.355      & 0.375      \\ \hline
\textbf{BOR12}   & 114         & 5           & 22          & 0.958      & 0.838      & 0.894      \\ \hline
\textbf{BOR17}   & 66          & 106         & 182         & 0.384      & 0.266      & 0.314      \\ \hline
\textbf{Total}   & 974         & 230         & 335         & 0.809      & 0.744      & 0.775      \\ \hline
\end{tabular}
\caption{Detailed per video results of T2001b. Here, we use S+BT for training our model. We report the combined results for both gradual and sharp transitions. We show the true positives (TP), false positives (FP), false negatives (FN), precision (P), recall (R) and F-measure (F).}
\label{2001b_01}
\end{table}


\begin{table*}[h]
\centering

\begin{tabular}{|c|c|c|c|c|c|c|c|c|c|c|c|c|c|c|}
\hline
 & \multicolumn{6}{c|}{\textbf{Gradual}} & \multicolumn{6}{c|}{\textbf{Sharp}}  \\ \hline
\textbf{Video}  & \textbf{\#T} & \textbf{TP} & \textbf{FP} & \textbf{FN} & \textbf{P} & \textbf{R} & \textbf{F} & \textbf{\#T} & \textbf{TP} & \textbf{FP} & \textbf{FN} & \textbf{P} & \textbf{R} & \textbf{F} \\ \hline
\textbf{BOR10\_001} & 11            & 11          & 0           & 0           & 1          & 1          & 1          & 0             & 0           & 0           & 0           & -          & -          & -          \\ \hline
\textbf{BOR10\_002} & 11            & 9           & 0           & 2           & 1          & 0.818      & 0.9        & 0             & 0           & 0           & 0           & -          & -          & -          \\ \hline
\textbf{NAD57}      & 25            & 22          & 1           & 3           & 0.957      & 0.88       & 0.917      & 45            & 45          & 4           & 0           & 0.918      & 1          & 0.957      \\ \hline
\textbf{NAD58}      & 44            & 37          & 0           & 7           & 1          & 0.841      & 0.914      & 40            & 33          & 0           & 7           & 1          & 0.825      & 0.904      \\ \hline
\textbf{anni001}    & 8             & 6           & 0           & 2           & 1          & 0.75       & 0.857      & 0             & 0           & 1           & 0           & 0          & -          & -          \\ \hline
\textbf{anni005}    & 27            & 26          & 2           & 1           & 0.929      & 0.963      & 0.945      & 39            & 36          & 13          & 3           & 0.735      & 0.923      & 0.818      \\
\hline
\end{tabular}
\caption{Detailed per video results of T2001a. Here, we use S+r+BT for training our model. We report the results for both gradual and sharp transitions. For each class we show the number of transitions (\#T), true positives (TP), false positives (FP), false negatives (FN), precision (P), recall (R) and F-measure (F).}
\label{2001a_06}
\end{table*}

\begin{table*}[h]
\centering

\begin{tabular}{|c|c|c|c|c|c|c|c|c|c|c|c|c|c|c|}
\hline
 & \multicolumn{6}{c|}{\textbf{Gradual}} & \multicolumn{6}{c|}{\textbf{Sharp}}  \\ \hline
\textbf{Video}  & \textbf{\#T} & \textbf{TP} & \textbf{FP} & \textbf{FN} & \textbf{P} & \textbf{R} & \textbf{F} & \textbf{\#T} & \textbf{TP} & \textbf{FP} & \textbf{FN} & \textbf{P} & \textbf{R} & \textbf{F} \\ \hline
\textbf{BOR10\_001} & 11           & 11          & 0           & 0           & 1          & 1          & 1          & 0            & 0           & 0           & 0           & -          & -          & -          \\ \hline
\textbf{BOR10\_002} & 11           & 10          & 0           & 1           & 1          & 0.909      & 0.952      & 0            & 0           & 0           & 0           & -          & -          & -          \\ \hline
\textbf{NAD57}      & 25           & 21          & 2           & 4           & 0.913      & 0.84       & 0.875      & 45           & 45          & 1           & 0           & 0.978      & 1          & 0.989      \\ \hline
\textbf{NAD58}      & 44           & 39          & 0           & 5           & 1          & 0.886      & 0.94       & 40           & 35          & 0           & 5           & 1          & 0.875      & 0.933      \\ \hline
\textbf{anni001}    & 8            & 6           & 0           & 2           & 1          & 0.75       & 0.857      & 0            & 0           & 0           & 0           & -          & -          & -          \\ \hline
\textbf{anni005}    & 27           & 27          & 1           & 0           & 0.964      & 1          & 0.982      & 39           & 35          & 5           & 4           & 0.875      & 0.897      & 0.886      \\
\hline
\end{tabular}
\caption{Detailed per video results of T2001a. Here, we use S+BT for training our model. We report the results for both gradual and sharp transitions. For each class we show the number of transitions (\#T), true positives (TP), false positives (FP), false negatives (FN), precision (P), recall (R) and F-measure (F).}
\label{2001a_01}
\end{table*}


\begin{table}[h]
\centering

\begin{tabular}{|c|c|c|c|c|c|c|c|}
\hline
& \multicolumn{6}{c|}{\textbf{Gradual and Sharp}}  \\ \hline
\textbf{Video}  & \textbf{TP} & \textbf{FP} & \textbf{FN} & \textbf{P} & \textbf{R} & \textbf{F} \\ \hline
\textbf{01811a}     & 60          & 7           & 4           & 0.896      & 0.938      & 0.916      \\ \hline
\textbf{6011}       & 40          & 96          & 81          & 0.294      & 0.331      & 0.311      \\ \hline
\textbf{8024}       & 85          & 22          & 21          & 0.794      & 0.802      & 0.798      \\ \hline
\textbf{8386}       & 113         & 10          & 5           & 0.919      & 0.958      & 0.938      \\ \hline
\textbf{8401}       & 26          & 5           & 5           & 0.839      & 0.839      & 0.839      \\ \hline
\textbf{10558a}     & 122         & 1           & 8           & 0.992      & 0.938      & 0.964      \\
\hline
\end{tabular}
\caption{Detailed per video results of T2002. Here, we use S+r+BT for training our model. We report the combined results for both gradual and sharp transitions. We show the true positives (TP), false positives (FP), false negatives (FN), precision (P), recall (R) and F-measure (F).}
\label{2002_07}
\end{table}

\begin{table}[h]
\centering

\begin{tabular}{|c|c|c|c|c|c|c|c|}
\hline
& \multicolumn{6}{c|}{\textbf{Gradual and Sharp}}  \\ \hline
\textbf{Video}  & \textbf{TP} & \textbf{FP} & \textbf{FN} & \textbf{P} & \textbf{R} & \textbf{F} \\ \hline
\textbf{01811a}       & 60          & 7           & 4           & 0.896      & 0.938      & 0.916      \\ \hline
\textbf{6011}        & 39          & 96          & 82          & 0.289      & 0.322      & 0.305      \\ \hline
\textbf{8024}        & 96          & 29          & 10          & 0.768      & 0.906      & 0.831      \\ \hline
\textbf{8386}        & 114         & 5           & 4           & 0.958      & 0.966      & 0.962      \\ \hline
\textbf{8401}         & 30          & 8           & 1           & 0.789      & 0.968      & 0.87       \\ \hline
\textbf{10558a}      & 125         & 1           & 5           & 0.992      & 0.962      & 0.977      \\
\hline
\end{tabular}
\caption{Detailed per video results of T2002. Here, we use S+BT for training our model. We report the combined results for both gradual and sharp transitions. We show the true positives (TP), false positives (FP), false negatives (FN), precision (P), recall (R) and F-measure (F).}
\label{2002_01}
\end{table}


\begin{table*}[h]
\centering

\begin{tabular}{|c|c|c|c|c|c|c|c|c|c|c|c|c|c|c|}
\hline
 & \multicolumn{6}{c|}{\textbf{Gradual}} & \multicolumn{6}{c|}{\textbf{Sharp}}  \\ \hline
\textbf{Video}     & \textbf{\#T} & \textbf{TP} & \textbf{FP} & \textbf{FN} & \textbf{P} & \textbf{R} & \textbf{F} & \textbf{\#T} & \textbf{TP} & \textbf{FP} & \textbf{FN} & \textbf{P} & \textbf{R} & \textbf{F} \\ \hline
\textbf{203\_CNN} & 171           & 134         & 44          & 37          & 0.753      & 0.784      & 0.768      & 280           & 228         & 13          & 52          & 0.946      & 0.814      & 0.875      \\ \hline
\textbf{222\_CNN} & 101           & 74          & 5           & 27          & 0.937      & 0.733      & 0.822      & 309           & 273         & 11          & 36          & 0.961      & 0.883      & 0.921      \\ \hline
\textbf{224\_ABC} & 131           & 108         & 10          & 23          & 0.915      & 0.824      & 0.867      & 296           & 281         & 13          & 15          & 0.956      & 0.949      & 0.953      \\ \hline
\textbf{412\_ABC} & 137           & 115         & 6           & 22          & 0.95       & 0.839      & 0.891      & 345           & 323         & 17          & 22          & 0.95       & 0.936      & 0.943      \\ \hline
\textbf{425\_ABC} & 180           & 161         & 12          & 19          & 0.931      & 0.894      & 0.912      & 295           & 266         & 11          & 29          & 0.96       & 0.902      & 0.93       \\ \hline
\textbf{515\_CNN} & 131           & 89          & 11          & 42          & 0.89       & 0.679      & 0.771      & 283           & 265         & 17          & 18          & 0.94       & 0.936      & 0.938      \\
\hline
\end{tabular}
\caption{Detailed per video results of T2003. Here, we use S+r+BT for training our model. We report the results for both gradual and sharp transitions. For each class we show the number of transitions (\#T), true positives (TP), false positives (FP), false negatives (FN), precision (P), recall (R) and F-measure (F).}
\label{2003_07}
\end{table*}

\begin{table*}[h]
\centering

\begin{tabular}{|c|c|c|c|c|c|c|c|c|c|c|c|c|c|c|}
\hline
 & \multicolumn{6}{c|}{\textbf{Gradual}} & \multicolumn{6}{c|}{\textbf{Sharp}}  \\ \hline
\textbf{Video}     & \textbf{\#T} & \textbf{TP} & \textbf{FP} & \textbf{FN} & \textbf{P} & \textbf{R} & \textbf{F} & \textbf{\#T} & \textbf{TP} & \textbf{FP} & \textbf{FN} & \textbf{P} & \textbf{R} & \textbf{F} \\ \hline
\textbf{203\_CNN} & 171          & 143         & 57          & 28          & 0.715      & 0.836      & 0.771      & 280          & 230         & 9           & 50          & 0.962      & 0.821      & 0.886      \\ \hline
\textbf{222\_CNN} & 101          & 80          & 24          & 21          & 0.769      & 0.792      & 0.78       & 309          & 275         & 11          & 34          & 0.962      & 0.89       & 0.924      \\ \hline
\textbf{224\_ABC} & 131          & 116         & 14          & 15          & 0.892      & 0.885      & 0.889      & 296          & 282         & 8           & 14          & 0.972      & 0.953      & 0.962      \\ \hline
\textbf{412\_ABC} & 137          & 122         & 11          & 15          & 0.917      & 0.891      & 0.904      & 345          & 323         & 11          & 22          & 0.967      & 0.936      & 0.951      \\ \hline
\textbf{425\_ABC} & 180          & 170         & 28          & 10          & 0.859      & 0.944      & 0.899      & 295          & 265         & 12          & 30          & 0.957      & 0.898      & 0.927      \\ \hline
\textbf{515\_CNN} & 131          & 105         & 16          & 26          & 0.868      & 0.802      & 0.833      & 283          & 259         & 15          & 24          & 0.945      & 0.915      & 0.93       \\
\hline
\end{tabular}
\caption{Detailed per video results of T2003. Here, we use S+BT for training our model. We report the results for both gradual and sharp transitions. For each class we show the number of transitions (\#T), true positives (TP), false positives (FP), false negatives (FN), precision (P), recall (R) and F-measure (F).}
\label{2003_01}
\end{table*}


\begin{table*}[h]
\centering

\begin{tabular}{|c|c|c|c|c|c|c|c|c|c|c|c|c|c|c|}
\hline
 & \multicolumn{6}{c|}{\textbf{Gradual}} & \multicolumn{6}{c|}{\textbf{Sharp}}  \\ \hline
\textbf{Video}      & \textbf{\#T} & \textbf{TP} & \textbf{FP} & \textbf{FN} & \textbf{P} & \textbf{R} & \textbf{F} & \textbf{\#T} & \textbf{TP} & \textbf{FP} & \textbf{FN} & \textbf{P} & \textbf{R} & \textbf{F} \\ \hline
\textbf{1004\_ABCa} & 203           & 166         & 13          & 37          & 0.927      & 0.818      & 0.869      & 224           & 213         & 22          & 11          & 0.906      & 0.951      & 0.928      \\ \hline
\textbf{1012\_CNNa} & 170           & 136         & 13          & 34          & 0.913      & 0.8        & 0.853      & 215           & 194         & 15          & 21          & 0.928      & 0.902      & 0.915      \\ \hline
\textbf{1016\_CNNa} & 150           & 119         & 9           & 31          & 0.93       & 0.793      & 0.856      & 242           & 214         & 13          & 28          & 0.943      & 0.884      & 0.913      \\ \hline
\textbf{1021\_ABCa} & 175           & 154         & 13          & 21          & 0.922      & 0.88       & 0.901      & 240           & 230         & 18          & 10          & 0.927      & 0.958      & 0.943      \\ \hline
\textbf{1101\_CNNa} & 204           & 172         & 20          & 32          & 0.896      & 0.843      & 0.869      & 191           & 187         & 11          & 4           & 0.944      & 0.979      & 0.961      \\ \hline
\textbf{1109\_ABCa} & 170           & 151         & 10          & 19          & 0.938      & 0.888      & 0.912      & 257           & 246         & 15          & 11          & 0.943      & 0.957      & 0.95       \\
\hline
\end{tabular}
\caption{Detailed per video results of T2004. Here, we use S+r+BT for training our model. We report the results for both gradual and sharp transitions. For each class we show the number of transitions (\#T), true positives (TP), false positives (FP), false negatives (FN), precision (P), recall (R) and F-measure (F).}
\label{2004_07}
\end{table*}

\begin{table*}[h]
\centering

\begin{tabular}{|c|c|c|c|c|c|c|c|c|c|c|c|c|c|c|}
\hline
 & \multicolumn{6}{c|}{\textbf{Gradual}} & \multicolumn{6}{c|}{\textbf{Sharp}}  \\ \hline
\textbf{Video}      & \textbf{\#T} & \textbf{TP} & \textbf{FP} & \textbf{FN} & \textbf{P} & \textbf{R} & \textbf{F} & \textbf{\#T} & \textbf{TP} & \textbf{FP} & \textbf{FN} & \textbf{P} & \textbf{R} & \textbf{F} \\ \hline
\textbf{1004\_ABCa} & 203           & 177         & 9           & 26          & 0.952      & 0.872      & 0.91       & 224           & 209         & 18          & 15          & 0.921      & 0.933      & 0.927      \\ \hline
\textbf{1012\_CNNa} & 170           & 149         & 23          & 21          & 0.866      & 0.876      & 0.871      & 215           & 191         & 13          & 24          & 0.936      & 0.888      & 0.912      \\ \hline
\textbf{1016\_CNNa} & 150           & 122         & 13          & 28          & 0.904      & 0.813      & 0.856      & 242           & 211         & 12          & 31          & 0.946      & 0.872      & 0.908      \\ \hline
\textbf{1021\_ABCa} & 175           & 154         & 22          & 21          & 0.875      & 0.88       & 0.877      & 240           & 227         & 14          & 13          & 0.942      & 0.946      & 0.944      \\ \hline
\textbf{1101\_CNNa} & 204           & 187         & 13          & 17          & 0.935      & 0.917      & 0.926      & 191           & 180         & 12          & 11          & 0.938      & 0.942      & 0.94       \\ \hline
\textbf{1109\_ABCa} & 170           & 159         & 12          & 11          & 0.93       & 0.935      & 0.933      & 257           & 241         & 11          & 16          & 0.956      & 0.938      & 0.947      \\
\hline
\end{tabular}
\caption{Detailed per video results of T2004. Here, we use S+BT for training our model. We report the results for both gradual and sharp transitions. For each class we show the number of transitions (\#T), true positives (TP), false positives (FP), false negatives (FN), precision (P), recall (R) and F-measure (F).}
\label{2004_01}
\end{table*}


\begin{table*}[h]
\centering

\begin{tabular}{|c|c|c|c|c|c|c|c|c|c|c|c|c|c|c|}
\hline
 & \multicolumn{6}{c|}{\textbf{Gradual}} & \multicolumn{6}{c|}{\textbf{Sharp}}  \\ \hline
\textbf{Video}                & \textbf{\#T} & \textbf{TP} & \textbf{FP} & \textbf{FN} & \textbf{P} & \textbf{R} & \textbf{F} & \textbf{\#T} & \textbf{TP} & \textbf{FP} & \textbf{FN} & \textbf{P} & \textbf{R} & \textbf{F} \\ \hline
\textbf{LNA}          & 198           & 147         & 11          & 51          & 0.93       & 0.742      & 0.826      & 45            & 31          & 26          & 14          & 0.544      & 0.689      & 0.608      \\ \hline
\textbf{NFC}     & 77            & 57          & 11          & 20          & 0.838      & 0.74       & 0.786      & 121           & 115         & 8           & 6           & 0.935      & 0.95       & 0.943      \\ \hline
\textbf{NEC}     & 94            & 88          & 3           & 6           & 0.967      & 0.936      & 0.951      & 74            & 57          & 5           & 17          & 0.919      & 0.77       & 0.838      \\ \hline
\textbf{HNA}         & 124           & 107         & 4           & 17          & 0.964      & 0.863      & 0.911      & 24            & 21          & 8           & 3           & 0.724      & 0.875      & 0.792      \\ \hline
\textbf{3PGC} & 228           & 171         & 32          & 57          & 0.842      & 0.75       & 0.794      & 132           & 112         & 48          & 20          & 0.7        & 0.848      & 0.767      \\ \hline
\textbf{CLE}       & 123           & 98          & 22          & 25          & 0.817      & 0.797      & 0.807      & 244           & 236         & 33          & 8           & 0.877      & 0.967      & 0.92       \\
\hline
\end{tabular}
\caption{Detailed per video results of T2006. Here, we use S+r+BT for training our model. We report the results for both gradual and sharp transitions. For each class we show the number of transitions (\#T), true positives (TP), false positives (FP), false negatives (FN), precision (P), recall (R) and F-measure (F).}
\label{2006_08}
\end{table*}

\begin{table*}[h]
\centering

\begin{tabular}{|c|c|c|c|c|c|c|c|c|c|c|c|c|c|c|}
\hline
 & \multicolumn{6}{c|}{\textbf{Gradual}} & \multicolumn{6}{c|}{\textbf{Sharp}}  \\ \hline
\textbf{Video}                & \textbf{\#T} & \textbf{TP} & \textbf{FP} & \textbf{FN} & \textbf{P} & \textbf{R} & \textbf{F} & \textbf{\#T} & \textbf{TP} & \textbf{FP} & \textbf{FN} & \textbf{P} & \textbf{R} & \textbf{F} \\ \hline
\textbf{LNA}          & 198           & 150         & 16          & 48          & 0.904      & 0.758      & 0.824      & 45            & 39          & 31          & 6           & 0.557      & 0.867      & 0.678      \\ \hline
\textbf{NFC}     & 77            & 56          & 22          & 21          & 0.718      & 0.727      & 0.723      & 121           & 115         & 8           & 6           & 0.935      & 0.95       & 0.943      \\ \hline
\textbf{NEC}     & 94            & 81          & 8           & 13          & 0.91       & 0.862      & 0.885      & 74            & 46          & 6           & 28          & 0.885      & 0.622      & 0.73       \\ \hline
\textbf{HNA}         & 124           & 113         & 33          & 11          & 0.774      & 0.911      & 0.837      & 24            & 22          & 6           & 2           & 0.786      & 0.917      & 0.846      \\ \hline
\textbf{3PGC} & 228           & 168         & 38          & 60          & 0.816      & 0.737      & 0.774      & 132           & 105         & 41          & 27          & 0.719      & 0.795      & 0.755      \\ \hline
\textbf{CLE}       & 123           & 110         & 28          & 13          & 0.797      & 0.894      & 0.843      & 244           & 223         & 14          & 21          & 0.941      & 0.914      & 0.927      \\
\hline
\end{tabular}
\caption{Detailed per video results of T2006. Here, we use S+BT for training our model. We report the results for both gradual and sharp transitions. For each class we show the number of transitions (\#T), true positives (TP), false positives (FP), false negatives (FN), precision (P), recall (R) and F-measure (F).}
\label{2006_01}
\end{table*}


\begin{table*}[h]
\centering

\begin{tabular}{|c|c|c|c|c|c|c|c|c|c|c|c|c|c|c|}
\hline
 & \multicolumn{6}{c|}{\textbf{Gradual}} & \multicolumn{6}{c|}{\textbf{Sharp}}  \\ \hline
\textbf{Video} & \textbf{\#T} & \textbf{TP} & \textbf{FP} & \textbf{FN} & \textbf{P} & \textbf{R} & \textbf{F} & \textbf{\#T} & \textbf{TP} & \textbf{FP} & \textbf{FN} & \textbf{P} & \textbf{R} & \textbf{F} \\ \hline
\textbf{BG\_11362} & 4             & 2           & 2           & 2           & 0.5        & 0.5        & 0.5        & 104           & 95          & 14          & 9           & 0.872      & 0.913      & 0.892      \\ \hline
\textbf{BG\_14213} & 61            & 49          & 0           & 12          & 1          & 0.803      & 0.891      & 106           & 106         & 3           & 0           & 0.972      & 1          & 0.986      \\ \hline
\textbf{BG\_2408}  & 20            & 17          & 4           & 3           & 0.81       & 0.85       & 0.829      & 101           & 100         & 5           & 1           & 0.952      & 0.99       & 0.971      \\ \hline
\textbf{BG\_34901} & 16            & 9           & 4           & 7           & 0.692      & 0.562      & 0.621      & 224           & 215         & 4           & 9           & 0.982      & 0.96       & 0.971      \\ \hline
\textbf{BG\_35050} & 4             & 1           & 0           & 3           & 1          & 0.25       & 0.4        & 98            & 98          & 0           & 0           & 1          & 1          & 1          \\ \hline
\textbf{BG\_35187} & 23            & 19          & 3           & 4           & 0.864      & 0.826      & 0.844      & 135           & 125         & 2           & 10          & 0.984      & 0.926      & 0.954      \\
\hline
\end{tabular}
\caption{Detailed per video results of T2007. Here, we use S+r+BT for training our model. We report the results for both gradual and sharp transitions. For each class we show the number of transitions (\#T), true positives (TP), false positives (FP), false negatives (FN), precision (P), recall (R) and F-measure (F).}
\label{2007_06}
\end{table*}

\begin{table*}[h]
\centering

\begin{tabular}{|c|c|c|c|c|c|c|c|c|c|c|c|c|c|c|}
\hline
 & \multicolumn{6}{c|}{\textbf{Gradual}} & \multicolumn{6}{c|}{\textbf{Sharp}}  \\ \hline
\textbf{Video} & \textbf{\#T} & \textbf{TP} & \textbf{FP} & \textbf{FN} & \textbf{P} & \textbf{R} & \textbf{F} & \textbf{\#T} & \textbf{TP} & \textbf{FP} & \textbf{FN} & \textbf{P} & \textbf{R} & \textbf{F} \\ \hline
\textbf{BG\_11362} & 4             & 0           & 2           & 4           & 0          & 0          & -        & 104           & 81          & 13          & 23          & 0.862      & 0.779      & 0.818      \\ \hline
\textbf{BG\_14213} & 61            & 45          & 2           & 16          & 0.957      & 0.738      & 0.833      & 106           & 106         & 3           & 0           & 0.972      & 1          & 0.986      \\ \hline
\textbf{BG\_2408}  & 20            & 18          & 5           & 2           & 0.783      & 0.9        & 0.837      & 101           & 100         & 7           & 1           & 0.935      & 0.99       & 0.962      \\ \hline
\textbf{BG\_34901} & 16            & 8           & 2           & 8           & 0.8        & 0.5        & 0.615      & 224           & 219         & 6           & 5           & 0.973      & 0.978      & 0.976      \\ \hline
\textbf{BG\_35050} & 4             & 0           & 1           & 4           & 0          & 0          & -        & 98            & 98          & 0           & 0           & 1          & 1          & 1          \\ \hline
\textbf{BG\_35187} & 23            & 19          & 1           & 4           & 0.95       & 0.826      & 0.884      & 135           & 125         & 3           & 10          & 0.977      & 0.926      & 0.951      \\
\hline
\end{tabular}
\caption{Detailed per video results of T2007. Here, we use S+BT for training our model. We report the results for both gradual and sharp transitions. For each class we show the number of transitions (\#T), true positives (TP), false positives (FP), false negatives (FN), precision (P), recall (R) and F-measure (F).}
\label{2007_01}
\end{table*}

\begin{table*}[h]
\small
\centering
\begin{tabular}{|c|c|c|c|c|c|c|c|}
     \hline      
				\textbf{Video} & \textbf{\#T} & \textbf{TP} & \textbf{FP} & \textbf{FN} & \textbf{P} & \textbf{R} & \textbf{F}  \\
			\hline 
				 V1                           & 80	& 66	& 12	& 14	& 0.846	& 0.825	& 0.835 \\ 
				 V2                           &146	&123	&8	&23	&0.939	&0.842	&0.888 \\ 
				 V3                           &112	&101	&1	&11	&0.99	&0.902	&0.944 \\
				 V4                           &60	&59	&2	&1	&0.967	&0.983	&0.975  \\ 
				 V5                           &104	&101	&3	&3	&0.971	&0.971	&0.971 \\ 
				 V6                           &54	&51	&4	&3	&0.927	&0.944	&0.936   \\
				 V7                           &109	&105	&3	&4	&0.972	&0.963	&0.968  \\ 
				 V8                           &196	&172	&5	&24	&0.972	&0.878	&0.922  \\ 
				 V9                           &61	&60	&1	&1	&0.984	&0.984	&0.984  \\
				 V10                          &63	&59	&0	&4	&1	&0.937	&0.967   \\				
			\hline
			 Overall                        &985	&897	&39	&88	&0.958	&0.911	&0.934       \\
			\hline
    \end{tabular}
\caption{\textit{Detailed per video results of the RAI dataset. Here, we use S+BT for training our model. We report the combined transition results. For each video we show the total number of transitions (\#T), true positives (TP), false positives (FP), false negatives (FN), precision (P), recall (R) and F-measure (F).}}
\label{tab:RAI}
\end{table*}

\begin{table*}[]
\centering

\begin{tabular}{|c|c|c|c|c|c|c|}
\hline

\textbf{Batch size} & \textbf{Starting Time} & \textbf{End Time} & \textbf{\# Seconds} & \textbf{Memory} & \textbf{\# Iterations} & \textbf{Faster than real time by} \\ \hline
\textbf{1}          & 19:08:23               & 19:11:26          & 183                 & 69413912        & 6394                   & 11.18076503                       \\ \hline
\textbf{1}          & 16:13:03               & 16:16:06          & 184                 & 69413912        & 6394                   & 11.12                             \\ \hline
\textbf{10}         & 21:16:31               & 21:18:37          & 125                 & 694139048       & 640                    & 16.36864                          \\ \hline
\textbf{10}         & 21:20:45               & 21:22:53          & 128                 & 694139048       & 640                    & 15.985                            \\ \hline
\textbf{20}         & 14:55:16               & 14:57:13          & 118                 & 1388278088      & 320                    & 17.33966102                       \\ \hline
\textbf{20}         & 14:59:21               & 15:01:19          & 117                 & 1388278088      & 320                    & 17.48786325                       \\ \hline
\textbf{30}         & 21:06:39               & 21:08:30          & 112                 & 2082417128      & 214                    & 18.26857143                       \\
\hline
\end{tabular}
\caption{The processing time of our technique. We report detailed analysis of different batch sizes as input. The bigger the batch size, the less processing time is required. This, however, requires more GPU memory. Experiments shows that the processing speed gain from 10 to 100 batch size is not significant. That’s between 16-19.3 real-time speed up factor.}  
\label{tab:PT}
\end{table*}

\end{document}